\documentclass[preprint,review,12pt,authoryear]{elsarticle}

\usepackage[T1]{fontenc}
\usepackage{psfrag, graphicx,color}
\usepackage{amssymb,amsmath,bm}
\usepackage{units,booktabs,url,makecell,xcolor,multirow, diagbox}
\usepackage[hidelinks]{hyperref}
\usepackage{tikz}
\usetikzlibrary{calc,fadings,arrows,shapes,backgrounds,plotmarks}
\usepackage{other/cmd_tex}
\usepackage{wrapfig}
\usepackage{subcaption}
\usepackage[prefix]{nomencl}
\usepackage{etoolbox}

\definecolor{OrangeOwn}{rgb}{0.9,0.38,0}
\definecolor{PurpleOwn}{rgb}{0.36,0.23,0.61}

\makenomenclature
\renewcommand\nomgroup[1]{%
	\item[\bfseries
	\ifstrequal{#1}{A}{\hspace{-.01cm}List of Acronyms}{
	\ifstrequal{#1}{S}{\hspace{-.15cm}List of Symbols and Indices}{}}
]}

\newcommand{\wrt}{\text{w.r.t. }}

\journal{Computers \& Chemical Engineering}

\makeatletter
\def\ps@pprintTitle{%
	\let\@oddhead\@empty
	\let\@evenhead\@empty
	\let\@oddfoot\@empty
	\let\@evenfoot\@oddfoot
}
\makeatother

\begin{document}
\begin{frontmatter}
\title{Data-Based Design of Multi-Model Inferential Sensors}
\author[STUBA,cor1]{Martin Mojto}
\author[SLOVNAFT]{Karol \v Lubu\v sk\'y}
\author[STUBA]{Miroslav Fikar}
\author[STUBA]{Radoslav Paulen}
\cortext[cor1]{Tel.: +421 (0)2 5932 5349, Mail: martin.mojto@stuba.sk (M. Mojto)}
\address[STUBA]{Faculty of Chemical and Food Technology, Slovak University of Technology in Bratislava, Radlinsk\'eho 9, 812 37 Bratislava, Slovakia}
\address[SLOVNAFT]{Slovnaft, a.s., Vl\v cie hrdlo 1, 824 12 Bratislava, Slovakia}

\begin{abstract}
	This paper deals with the problem of inferential (soft) sensor design. The nonlinear character of industrial processes is usually the main limitation to designing simple linear inferential sensors with sufficient accuracy. In order to increase the inferential sensor predictive performance and yet to maintain its linear structure, multi-model inferential sensors represent a straightforward option. In this contribution, we propose two novel approaches for the design of multi-model inferential sensors aiming to mitigate some drawbacks of the state-of-the-art approaches. For a demonstration of the developed techniques, we design inferential sensors for a Vacuum Gasoil Hydrogenation unit, which is a real-world petrochemical refinery unit. The performance of the multi-model inferential sensor is compared against various single-model inferential sensors and the current (referential) inferential sensor used in the refinery. The results show substantial improvements over the state-of-the-art design techniques for single-/multi-model inferential sensors.
\end{abstract}

\end{frontmatter}

\printnomenclature
\newpage

\section{Introduction}\label{sec:introduction}
\nomenclature[A,01]{GF}{Gasoline Fraction}
\nomenclature[A,02]{HGO}{Hydrogenated Gasoil}
\nomenclature[A,03]{LASSO}{Least Absolute Shrinkage and Selection Operator}
\nomenclature[A,04]{MILP}{Mixed-Integer Linear Program}
\nomenclature[A,05]{MINLP}{Mixed-Integer NonLinear Program}
\nomenclature[A,06]{MIS}{Multi-Model Inferential Sensor}
\nomenclature[A,07]{NN}{Neural Network}
\nomenclature[A,08]{OLSR}{Ordinary Least Squares Regression}
\nomenclature[A,09]{PCR}{Principal Component Regression}
\nomenclature[A,10]{PLSR}{Partial Least Squares Regression}
\nomenclature[A,11]{RMSE}{Root Mean Squared Error}
\nomenclature[A,12]{SAE}{Sum of Absolute Errors}
\nomenclature[A,13]{SIS}{Single-Model Inferential Sensor}
\nomenclature[A,14]{SS}{Subset Selection}
\nomenclature[A,15]{SSE}{Sum of Squared Errors}
\nomenclature[A,16]{SVM}{Support Vector Machine}
\nomenclature[A,17]{VGH}{Vacuum Gasoil Hydrogenation}
\nomenclature[S,01]{$\bm{a}$}{Vector of inferential sensor parameters, $a\in\R^{n_{\text{p}}}$}
\nomenclature[S,02]{$a_0$}{Constant inferential sensor off-set (bias)}
\nomenclature[S,03]{$\alpha$}{Weighting parameter for normal vector of the separation plane}
\nomenclature[S,04]{$\beta$}{Weighting parameter for vector of the slack variables}
\nomenclature[S,05]{col}{Distillation column section}
\nomenclature[S,06]{$\mathcal D$}{Available dataset}
\nomenclature[S,07]{$\bm{e}$}{Vector of the slack variables, $\bm{e}\in\R^{n}$}
\nomenclature[S,08]{F}{Feed stream} 
\nomenclature[S,09]{$F$}{Flowrate} 
\nomenclature[S,10]{h}{Heating medium to a reboiler}
\nomenclature[S,11]{$H_v$}{Heat of vaporization}
\nomenclature[S,12]{$L$}{Liquid level}
\nomenclature[S,13]{$\lambda$}{Weighting parameter of LASSO}
\nomenclature[S,14]{$\bm{m}_i$}{Vector of input variables for $i^{\text{th}}$ measurement, $\bm{m}_i\in\R^{n_{\text{p}}}$}
\nomenclature[S,15]{$\mu$}{Mean value}
\nomenclature[S,16]{$n$}{Number of measurements}
\nomenclature[S,17]{$n_{\text{p}}$}{Number of input variables (candidates)} 
\nomenclature[S,18]{$n_{\text{pc}}$}{Number of principal components} 
\nomenclature[S,19]{$p$}{Pressure} 
\nomenclature[S,20]{$PCT$}{Pressure compensated temperature}
\nomenclature[S,21]{$R$}{Universal gas constant}
\nomenclature[S,22]{$\mathcal R$}{Convex polyhedra}
\nomenclature[S,23]{reb}{Reboiler}
\nomenclature[S,24]{rt}{Ratio}
\nomenclature[S,25]{$\mathcal S$}{Testing dataset}
\nomenclature[S,26]{$\sigma$}{Standard deviation}
\nomenclature[S,27]{t}{Top section (location)} 
\nomenclature[S,28]{$T$}{Termodynamic temperature} 
\nomenclature[S,29]{$\mathcal T$}{Training dataset}
\nomenclature[S,30]{vo}{Valve opening} 
\nomenclature[S,31]{vap}{Vapor phase}
\nomenclature[S,32]{$\bm{w}$}{Normal vector of the separation hyperplane, $w\in\R^{n_{\text{p}}}$}
\nomenclature[S,33]{$w_0$}{Constant separation hyperplane off-set (bias)}
\nomenclature[S,34]{$x$}{Concentration (composition)} 
\nomenclature[S,35]{$y$}{Output (desired) variable, $y\in\R$}
\nomenclature[S,36]{$\hat{y}$}{Estimated output (desired) variable, $\hat{y}\in\R$}
\nomenclature[S,37]{$\bm{z}$}{Vector of the binary variables, $\bm{z}\in\R^{n}$ or $\bm{z}\in\R^{n_\text{p}}$}

Increasing need for automation triggers a growing demand in the industry for accurate inferential (soft) sensors. The purpose of the inferential sensor is to estimate desired hard-to-measure variables (e.g., product concentration) using measurements from the online process sensors (e.g., temperatures, pressures, flow rates). The use of an inferential sensor represents a cheap alternative that can provide more frequent sensing of the desired variables compared to other ways of process monitoring, e.g., physical sensing devices or lab analysis.

In general, the production processes represent complex systems with many variables and interactions between these variables~\citep{santander_2022} and they usually exhibit nonlinear behavior resulting from the rich interactions of involved physical phenomena. One would conjecture that a nonlinear inferential sensor design is necessary. However, a typical industrial process is usually operated in some target operating range to achieve desired product specifications. Therefore, the nonlinear behavior of the process variable can often be neglected, and the linear inferential sensor can provide an accurate estimate of the desired variable. The advantage of a linear inferential sensor over some nonlinear counterpart lies foremost in lower maintenance expenses, higher transparency, possibility of physical insight, lower computational effort for sensor training, validation, online evaluation, and further calculations (e.g., for optimization and control). The later aspect can be significant mainly when the estimated variable is an input for an advanced process controller~\citep{botha_2021} or it is involved in a (complex) plant-wide optimization~\citep{ge_2017}.

The structure of the linear inferential sensor usually involves only a single model (hence, single-model inferential sensor (SIS)). According to the SIS design approach, one can distinguish between model-based~\citep{doraiswami_2014}, data-driven~\citep{king_2011, mojto_2021, sun_2021} and hybrid approaches~\citep{tahir_2019, zhuang_2022}. The complexity and size of industrial processes significantly limits the use of the model-based SIS modeling techniques. Therefore, the data-driven SIS approaches are predominantly used in industrial practice. The most popular advanced design techniques include Principal Components Regression (PCR)~\citep{kendall_1957}, Partial Least Squares regression (PLSR)~\citep{wold_1984, wold_2001}, the Least Absolute Shrinkage and Selection Operator (LASSO) method~\citep{santosa_1986, tibshirani_2011}, and the Subset Selection (SS) approaches~\citep{konno_2009, miyashiro_2015, takano_2020}. The usage of these techniques to provide an automatic modeling framework is presented in~\cite{sun_2021}. Moreover, a comprehensive analysis of the data-driven approaches can be found in~\cite{mojto_2021, sun_2022}. These contributions compare various performance criteria (RMSE, $R^2$, model complexity) of these approaches considering the real data from complex industrial case studies.

Multi-model linear inferential sensors bridge the gap between linear and nonlinear inferential sensors. To retain the advantages of the linear inferential sensor yet to increase the prediction performance, it is possible to consider multiple linear models within the sensor structure (hence, multi-model inferential sensor (MIS)). Each model within the MIS structure can explain a particular part of the operating region, e.g. a cluster of possible steady states. Therefore, these sensors find applications in complex industrial processes with multitude of operating modes~\citep{khatibisepehr_2012, jin_2015, wang_2021}.

The state-of-the-art MIS design consists of three sequential steps: (1) a priori labeling, (2) data classification, and (3) individual training of the inferential sensor models. In the first step, a modeler searches for an appropriate number of models and assigns tags to available data to distinguish the models (classes) discovered. The popular approach for a priori labeling is $k$-means clustering~\citep{forgy_1965}. A comparison of several other techniques for a priori labeling is shown in~\cite{lu_2014}. The classification step employs an appropriate data-based (machine-learning) approach to draw model-validity regions, i.e., the boundaries between the classes (models) that would later serve as switching conditions for using predictions from a particular model. A frequently used and well-known classification learner is Support Vector Machine (SVM)~\citep{boser_1992}. The method designs classification hyperplanes in the context studied in this work. Lastly, the constituent models of MIS are individually trained for each class by using a suitable regression technique~\citep{mojto_2021}. One of the recent examples of learning MIS is given in~\cite{bemporad_2022}, where piecewise linear regression is considered together with classification based on softmax regression and labeling by the $k$-means algorithm. A similar approach~\citep{ferrari_2002} uses a neural network as a classifier. While the state-of-the-art approaches train MIS effectively, there are still a few drawbacks hindering the overall potential of MIS. The first drawback is that the continuity at the switch between the different MIS models is not guaranteed. This can have a negative impact on the plant production. For example, a common issue might arise that advanced process controllers with MIS implemented might face stability issues because of inferential sensor discontinuity. The second drawback originates from the a priori labeling that is unaware --- like any other unsupervised learning approach --- of its impact on MIS prediction accuracy. It is, therefore, not likely that the optimal allocation of the model-validity regions is achieved.

In this paper, we propose two novel approaches that prune the MIS design of the negative effects of the drawbacks mentioned above. Firstly, both the proposed approaches ensure continuity when switching between the designed MIS models. This is achieved by merging classification and model training into one decision problem and training an SVM separation hyperplane to act as a switching boundary between the MIS models. Secondly, we propose an optimization-based labeling approach, effectively conducting all the sequential steps of the state-of-the-art MIS design procedure simultaneously. The performance of the proposed approaches is compared on the synthetic dataset from a model of pressure compensated temperature $PCT$ and subsequently on the industrial dataset from the Vacuum Gasoil Hydrogenation (VGH) unit. The VGH unit represents a crucial section within the oil refinery Slovnaft, a.s. in Bratislava, Slovakia. The performance of MIS is compared (a) with the reference SIS currently used in the refinery and (b) with SIS designed by Ordinary Least Squares Regression (OLSR), PCR, PLSR, LASSO, and SS. The conclusions in this study follow from the standard indicators, such as prediction accuracy (RMSE) and complexity (i.e., number of model input variables or principal components).

This paper is organized as follows. In Section~\ref{sec:prob_description}, the basic description of SIS and MIS is introduced. Section~\ref{sec:solution_approach} shows the development of a novel approach for the MIS design. Section~\ref{sec:case_studies} shows the design of SIS and MIS for case studies from the petrochemical industry. In Section~\ref{sec:discussion}, the results are discussed. Finally, in Section~\ref{sec:conclusions}, we summarize the main conclusions and remarks of this work.

\section{Problem Description}\label{sec:prob_description}
We aim at designing an inferential sensor for predicting a hard-to-measure process variable $y\in\R$ based on easy-to-measure process variables $\bm{m}\in\R^{n_\text{p}}$, which are also referred to as inputs of the sensor. A mathematical representation of the sensor reads as:
\begin{align}\label{eq:is}
	\hat{y} = f\left(\bm{m}\right),
\end{align}
where $\hat{y}\in\R$ is the prediction of $y$, which is also referred to as output of the sensor. We consider just scalar output but inclusion of multiple inputs is a straightforward extension of this work. We focus on static sensors as they are typically desired in industry and because they often represent the appropriate model structure. For example, in the petrochemical industry, a concentration of distillation products fits a nonlinear static thermodynamic function of process variables (e.g., temperatures and pressures).

\subsection{Design of a Single-Model Inferential Sensor}\label{sec:sis}
We consider a linear SIS that can be represented as~\citep{mojto_2021}:
\begin{align}\label{eq:sis}
 \hat{y} = \bm{m}^\intercal\bm{a} + a_0,
\end{align}
where $\bm{a}:=\left(a_1, a_2, \ldots, a_{n_{\text{p}}}\right)^\intercal\in\R^{n_\text{p}}$ represents the vector of sensor parameters and $a_0$ is a constant off-set (bias).

The SIS design is given by three subsequent stages~\citep{khatibisepehr_2013,botha_2021,mojto_2021}:
\begin{enumerate}
    \item \emph{Data pre-processing stage}. This usually includes standardization for the data (series of $y$ and $\bm m$) to involve zero mean and unit standard deviation, removal of systematic errors and outliers, and selection of the $n_\text{p}$ input candidates from the whole set of input variables (e.g., based on linear independence and correlation criteria).
    \item \emph{Training stage}. The training dataset (with a corresponding index set $\mathcal T$) is being considered in this step, which is a subset of the entire available dataset (index set $\mathcal D$). The model parameters are calculated based on a chosen fitting criterion. Sensor structure is selected directly through some measures to prevent overfitting (e.g., cross validation on a validation dataset) or using feature selection, or indirectly by including such measure in the calculation of the model parameters.
    \item \emph{Testing stage}. This optional step uses (previously unseen) testing dataset (index set $\mathcal S$; $\mathcal{S} := \mathcal{D} \setminus \mathcal{T}$). The purpose of this stage is, for example, to decide between several candidate inferential sensors trained in parallel according to the previous stages. The assessment measure is some accuracy criterion (e.g., root mean squared error (RMSE)).
\end{enumerate}

\subsubsection{Training Methods for SIS}\label{sec:sis_train}
We list here some methods of popular choice~\citep{khatibisepehr_2013,mojto_2021} for SIS training phase.
\begin{itemize}
\item Ordinary Least Squares Regression (OLSR)

The parameters are obtained by solving:
\begin{equation}\label{eq:ols}
	\min_{\bm{a},a_0} \frac{1}{2}\sum_{\forall i\in\mathcal T}(y_{i}-\bm{m}_{i}^\intercal \bm{a}-a_0)^2,
\end{equation}
where index $i$ stands for a measurement index such that $\text{card}(\mathcal T) > n_{\text{p}}$.

\item Principal Component Regression (PCR)~\citep{kendall_1957}

This statistical approach reduces the dimensionality of a dataset by linear transformation into a new orthogonal space of principal components ($n_\text{p} \rightarrow n_\text{pc}, n_\text{p} \geq n_\text{pc}$). The aim is to preserve most of the information content (variation) within the original dataset by using as few principal components as possible. The principal components are derived by singular value decomposition of a covariance matrix, and therefore, PCR is considered an unsupervised learning approach. However, it can be effectively combined with some of the regression approaches (e.g., OLSR) in order to design a linear SIS. 

\item Partial Least Squares Regression (PLSR) regression~\citep{wold_1984}

This statistical approach shares similar characteristics with PCR. The optimization problem~\eqref{eq:ols} is solved over a transformed input space of principal components. The designed principal components are derived by singular value decomposition of a cross-covariance matrix, and therefore, PLSR is considered a supervised learning approach, unlike PCR.

\item Least Absolute Shrinkage and Selection Operator (LASSO) regression~\citep{santosa_1986}

This method finds the sensor structure by solving:
\begin{equation}\label{eq:lasso}
	\min_{\bm{a},a_0} \ \frac{1}{2}\sum_{\forall i\in\mathcal T}(y_{i}-\bm{m}_{i}^\intercal \bm{a} - a_0)^2 + \lambda\|\bm{a}\|_1,
\end{equation}
where $\lambda$ is a weight between the accuracy of the model training and the model overfitting. The resulting model structure is trained using~\eqref{eq:ols} while the inputs corresponding to zero coefficients from the solution to~\eqref{eq:lasso} being discarded.

\item Subset Selection (SS)~\citep{konno_2009}
The SS approach seeks the best sensor with a given number of inputs. One solves:
\begin{subequations}\label{eq:SS_CV}
	\begin{align}
		&\min_{\substack{\bm{a}, a_0\\ \bm{z}\in \{0, 1\}^{n_\text{p}}}} \ \frac{1}{2}\sum_{i\in\mathcal T}\left(y_i - \bm{m}_i^\intercal\bm{a}-a_0\right)^2\\
		& \qquad \text{s.t. } -\bar az_k \leq \bm{a} \leq \bar az_k, \ \forall k\in\{1,\dots,n_\text{p}\},\\
		& \qquad \qquad \sum_{k=1}^{n_\text{p}} z_k = \tilde n_\text{p},
	\end{align}
\end{subequations}
where $\tilde n_\text{p}$ is the number of desired sensor inputs, $\bar a$ represents an upper bound on $\|\bm{a}\|_\infty$ and $\bm{z}$ denotes a vector with binary entries signifying the selection of the $k^\text{th}$ input into the sensor structure. A cross-validation approach is commonly used to determine best $\tilde n_\text{p}$, a final form of the model.

We note that the SS approach can be also enhanced to directly seek a trade-off between the simplicity and accuracy of the sensor. Here, a chosen cross-validation criterion can be used explicitly in the design~\citep{miyashiro_2015,takano_2020, mojto_2021}.
\end{itemize}

\paragraph{Illustrative example.} To motivate our study, we consider a problem of designing an inferential sensor for the pressure compensated temperature $PCT$. This variable is frequently used in low-pressure petrochemical distillation columns~\citep{pan_2019}. A combination of the Antoine and Clausius-Clapeyron equations forms the following mathematical representation~\citep{king_2011}:
\begin{equation}\label{eq:pct_model}
	\frac{1}{PCT} = \frac{R}{H_v}\ln\left(\frac{p}{p_{\text{ref}}}\right) + \frac{1}{T},
\end{equation}
where $H_v$ is the heat of vaporization, $R$ is the universal gas constant, $p_{\text{ref}}$ is the reference pressure, $p$ is the absolute pressure, and $T$ is the absolute temperature. 

The ground truth model of the $PCT$ is considered with $R = 8.3$\,J/mol/K, $H_v = 55,940.6$\,J/mol and $p_\text{ref} = 145.3$\,Pa over the operating region:
\begin{equation}\label{eq:pct_intervals}
	\begin{aligned}
		523.2\,\text{K} \leq &\,\,T\leq 573.2\,\text{K},\\
		0.4\,\text{Pa} \leq &\,\,p\leq 15\,\text{Pa},\\
		635.3\,\text{K} \leq &\,\,PCT\leq 1151.4\,\text{K}.
	\end{aligned}
\end{equation}
For sensor training and evaluation, all input variables ($p$, $T$, $PCT$) are scaled (normalized) to the interval $[0,1]$.

\begin{figure}
	\centering
	\begin{subfigure}[b]{.495\textwidth}
		\includegraphics[width=\linewidth]{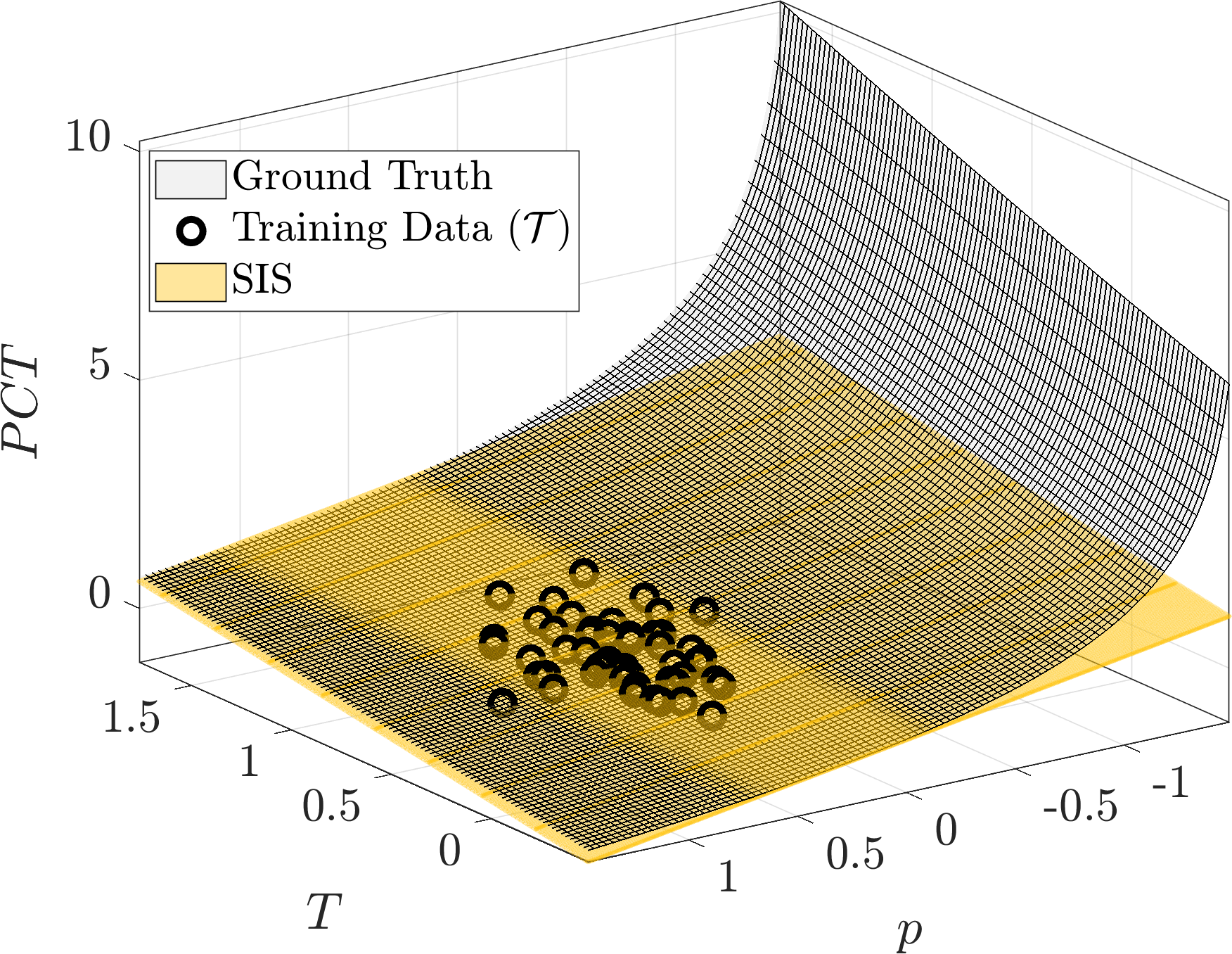}
		\caption{One distinct cluster (RMSE~($\mathcal T$)=0.028).}
		\label{fig:pct_SIS_1cls}
	\end{subfigure}
	\begin{subfigure}[b]{.495\textwidth}
		\includegraphics[width=\linewidth]{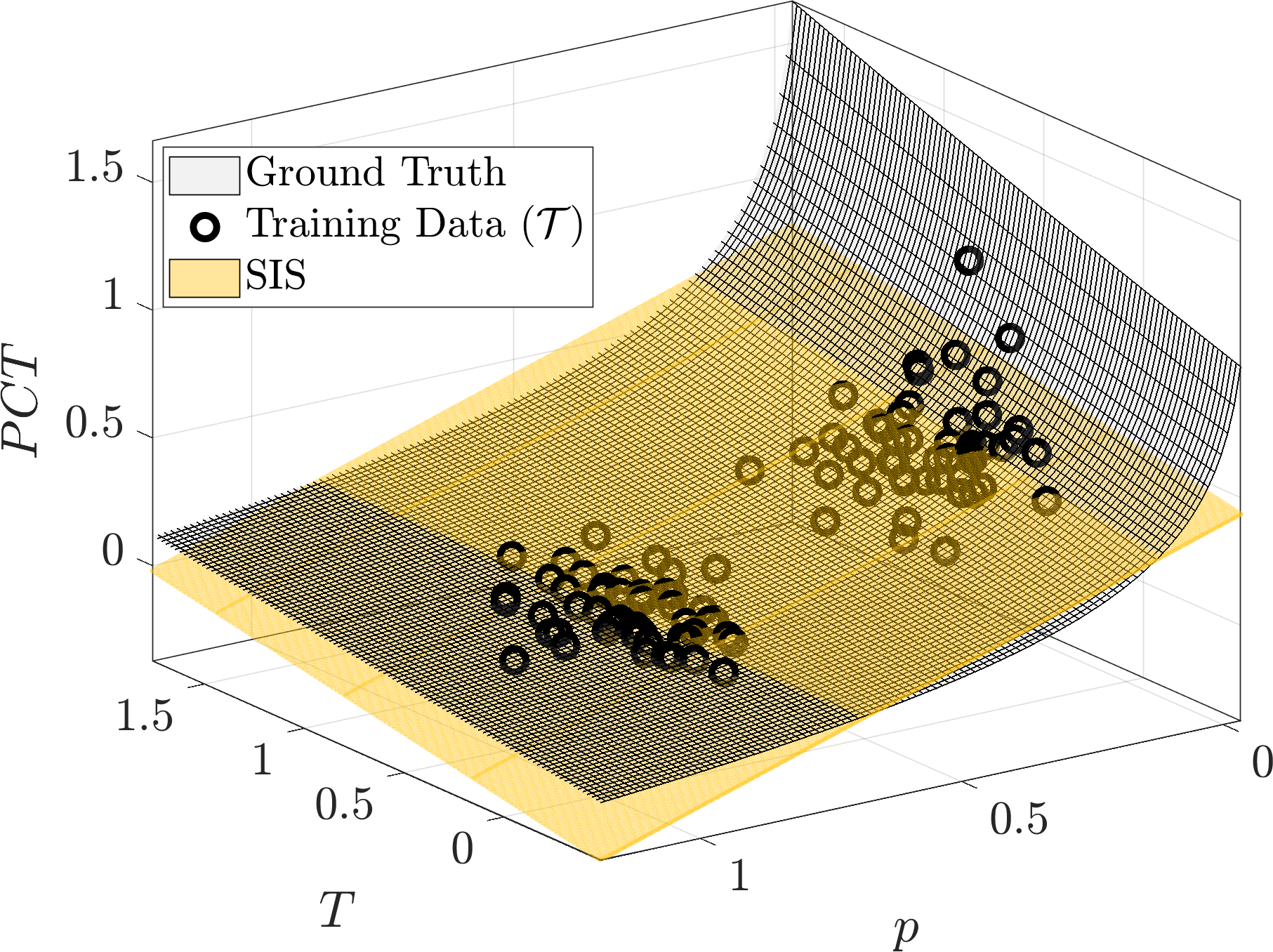}
		\caption{Two distinct clusters (RMSE~($\mathcal T$)=0.064).}
		\label{fig:pct_SIS_2cls}
	\end{subfigure}
	\caption{The ground truth model of $PCT$ with SIS designed on different datasets.}
	\label{fig:pct_SIS}
\end{figure}

The SIS performance is shown in Figure~\ref{fig:pct_SIS} on two training datasets, which simulates the process working in one (one data cluster in Figure~\ref{fig:pct_SIS_1cls}) and two distinct operating regimes (two data clusters in Figure~\ref{fig:pct_SIS_2cls}), respectively. Zero-mean Gaussian white noise with a standard deviation $\sigma_\text{noise} = 5\,\text{K}$ is added to the output data. The SIS accuracy, measured by the root mean squared error (RMSE) and tested on fresh data within the training regions, is significantly reduced (more than 2-fold deterioration) when the process runs in two operating regimes. This stems from inappropriateness of a single linear model to describe a nonlinear behavior of $PCT$.

\subsection{Multi-Model Inferential Sensor}\label{sec:mis}
Prediction capability of a linear inferential sensor can be improved when considering a multi-model sensor structure. The MIS formulation with two models can be written as follows~\citep{mojto_2022}:
\begin{equation}\label{eq:mis_def}
	\hat{y}=
	\begin{cases} 
		\bm{m}^\intercal \bm{a}_1 + a_{0,1}, & \text{if }\bm{m} \in \mathcal R_1,\\
		\bm{m}^\intercal \bm{a}_2 + a_{0,2}, & \text{if }\bm{m} \in \mathcal R_2,
	\end{cases}
\end{equation}
where $\bm{a}_1$ and $\bm{a}_2$ represent vectors of parameters of the first and second model, respectively, and $a_{0,1}$ and $a_{0,2}$ are constant off-sets of the first and second model, respectively. The regions of individual model validity are denoted as $\mathcal R_j$ and represent convex polyhedra such that $\mathcal R_1\bigcap\mathcal R_2=\emptyset$. Consideration of more than two models is possible in a similar setup.

Design of MIS follows a three stages as described for SIS. The design phase uses the following state-of-the-art workflow:
\begin{enumerate}
	\item \emph{Clustering for a priori labeling of the training dataset.} The labeling results from the dataset characteristics (e.g., distinction of operating points). An appropriate clustering approach (e.g., $k$-means clustering) can be used.
	\item \emph{Classifier training based on the labeled training dataset.} The purpose of the classifier is to determine the corresponding model class of a measurement point. In this paper, the support vector machines (SVM) approach is considered using linear separators~\citep{boser_1992}. The separators establish the polyhedral partitions $\mathcal R_j$ in~\eqref{eq:mis_def}.
	\item \emph{Training of the individual MIS models.} The individual MIS models for each class can be fitted using some of the SIS training methods (see in Section~\ref{sec:sis_train}).
\end{enumerate}

\begin{figure}
	\centering
	\includegraphics[width=\linewidth]{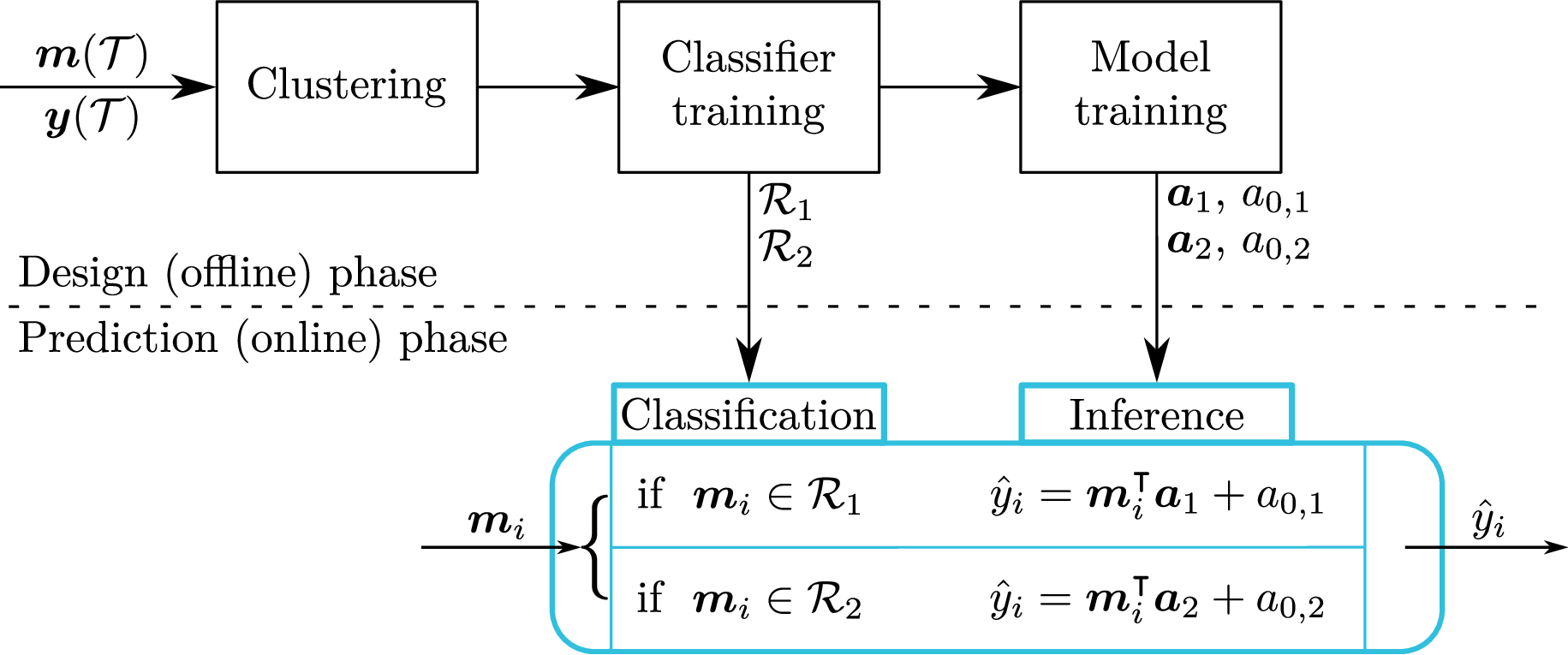}
	\caption{A simplified workflow of design and prediction phases of MIS.}
	\label{fig:mis_design}
\end{figure}

The design phase of MIS is illustrated in Figure~\ref{fig:mis_design} (upper part). This phase is performed offline using the available input and output training datasets ($\bm m(\mathcal T)$ and $\bm y(\mathcal T)$). Its purpose is to establish or update classification regions ($\mathcal R_1$ and $\mathcal R_2$) and model parameters ($\bm a_1$, $\bm a_2$, $a_{0,1}$, and $a_{0,2}$) required for the prediction of MIS, as depicted in the bottom section of Figure~\ref{fig:mis_design}. In contrast to the design phase, the prediction phase is conducted online, while each incoming (actual) measurement of input variables ($\bm m_i$) undergoes a classification to be assigned to one of the considered regions ($\mathcal R_1$ or $\mathcal R_2$). Subsequently, the actual prediction ($y_i$) is computed using the individual MIS model corresponding to the determined class (region). These calculations are performed within the time period represented by index $i$.

In the following, we abbreviate the sensor designed by this procedure as MIS$_\text{SotA}$ and we refer it to as a state-of-the-art approach although the presented procedure is our contribution. This is because there exist no consistent (agreed-upon) technique for the MIS design according to the best of authors' knowledge.

The MIS training assumes an a priori choice of the number of models within the MIS structure. Naturally, more the number of models considered, higher the susceptibility of the resulting sensor to overfitting. The number of models within the MIS structure is thus a tuning parameter of the scheme, whose value can be decide based on cross-validation.

\begin{figure}
	\centering
	\begin{subfigure}[b]{.495\textwidth}
		\includegraphics[width=\linewidth]{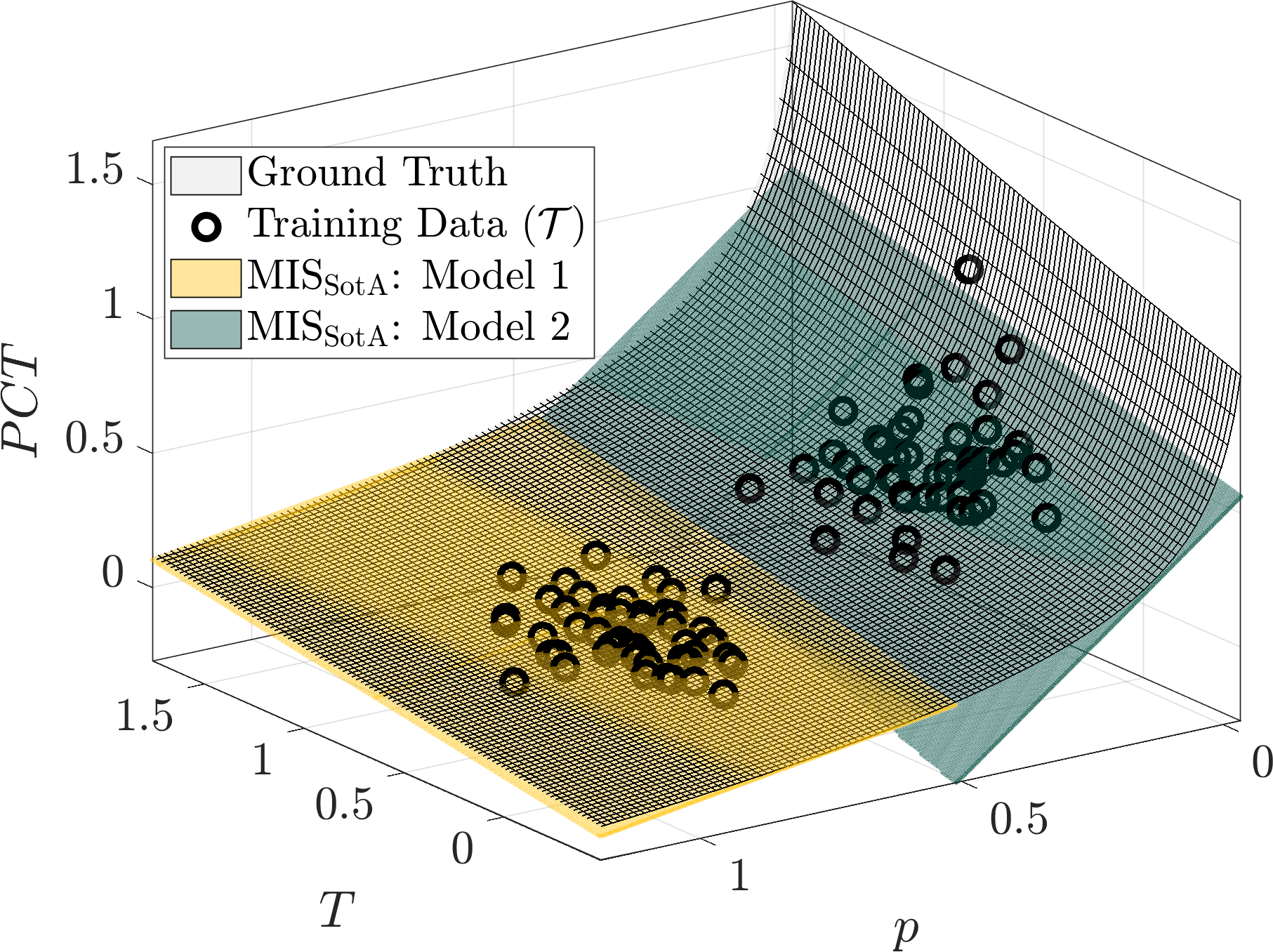}
		\caption{MIS$_\text{SotA}$ (RMSE~$(\mathcal T)=0.032$).}
		\label{fig:pct_MIS_SotA_2cls}
	\end{subfigure}
	\begin{subfigure}[b]{.495\textwidth}
		\includegraphics[width=\linewidth]{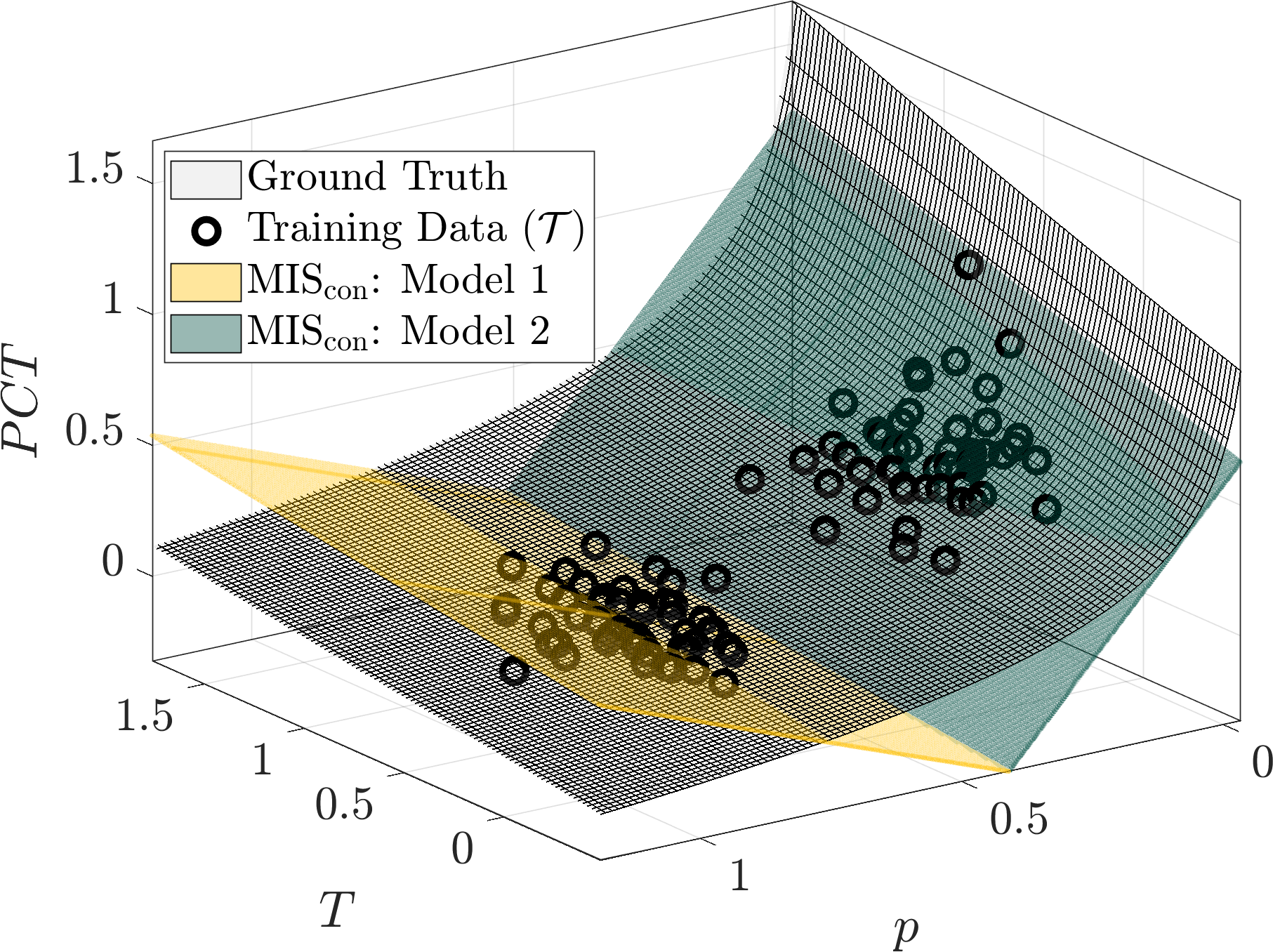}
		\caption{MIS$_\text{con}$ (RMSE~$(\mathcal T)=0.079$).}
		\label{fig:pct_MIS_con_2cls}
	\end{subfigure}
	\caption{The ground truth model of $PCT$ with MIS$_\text{SotA}$ and MIS$_\text{con}$ designed on the dataset with two distinct clusters.}
	\label{fig:pct_MIS_SotA_vs_MIS_con_2cls}\vspace{0cm}
\end{figure}

\paragraph{Illustrative example (continued)} The usage of the MIS$_\text{SotA}$ on the $PCT$ dataset is shown in Figure~\ref{fig:pct_MIS_SotA_2cls}. The designed models are presented with a yellow surface (Model 1) and a dark green surface (Model 2). The example considers a priori labeling by $k$-means clustering.

The advantage of using MIS is obvious, as its accuracy (RMSE~$(\mathcal T)=0.032$) significantly outperforms the best SIS (Figure~\ref{fig:pct_SIS_2cls}, RMSE~$(\mathcal T)=0.064$). This confirms that the MIS models can better explain the nonlinear behavior of $PCT$ compared to SIS. Furthermore, the structure of MIS is flexible as it can involve more models. There are two primary limitations (challenges) of MIS$_\text{SotA}$: (a) the designed models are not necessarily continuous, and (b) a priori labeling is unaware of its impact on the accuracy of the resulting inferential sensor. The first drawback can be seen in Figure~\ref{fig:pct_MIS_SotA_2cls}. There is a visible discrepancy between the designed models of MIS$_\text{SotA}$ at the intersection of the surfaces. This behavior can cause issues with the stability of the control strategy if the MIS is involved. 

A glimpse of the proposed solution in Figure~\ref{fig:pct_MIS_con_2cls} (approach MIS$_\text{con}$ will be introduced in detail in Section~\ref{sec:mis_con}) reveals that it is possible to achieve continuity when switching between the MIS models, yet potentially, at the expense of model accuracy. In the studied example, the discrepancy between ground truth and the designed models originates from the rotation of Model~1 (the yellow surface in Figure~\ref{fig:pct_MIS_con_2cls}) to achieve the desired continuity when switching to Model~2 (the dark green surface). The rotation can be reduced by putting more weight on accuracy and relaxing the continuity constraint when designing a continuous MIS (as discussed in Section~\ref{sec:mis_con}) or optimizing a priori labeling (as discussed in Section~\ref{sec:mis_con_lab}).

\section{Solution Approach}\label{sec:solution_approach}
We propose a novel approach for the MIS design consisting of two separate developments dealing with the limitations of the MIS$_\text{SotA}$ approach. The first part deals with the approach for the design of continuous MIS (referred to as a MIS$_\text{con}$ approach). The second part extends the design of MIS with optimized data labeling (referred to as a MIS$_\text{con,lab}$ approach).
 
\subsection{Design of a Continuous MIS}\label{sec:mis_con}
To deal with the limitation of discontinuity of MIS models, we propose a combination of the SVM-based classification of the data with the individual sensor training in the following optimization problem:
\begin{subequations}\label{eq:mis_con}
	\begin{align}\label{eq:mis_con_obj}
		&\min_{\substack{\bm{w}, w_0, \bm{e}\geq0\\\bm{a}_1, a_{0,1}, \bm{a}_2, a_{0,2}}}\text{SSE}_1 + \text{SSE}_2 + \alpha\|\bm{w}\|_2^2 + \beta\|\bm{e}\|_1\\
		\label{eq:mis_con_ctd_const_1}
		&\qquad\text{s.t. }\quad (2z_{i}-1)\left(\bm{m}_i^\intercal \bm{w} + w_{0}\right) \geq 1 - e_{i}, \quad \forall i \in \{1, 2, \ldots, n\},\\
		\label{eq:mis_con_ctd_const_2}
		&\qquad\quad\qquad \text{SSE}_1=\sum_{\forall i\in\mathcal T}z_{i}\left(y_i - \bm{m}_i^\intercal \bm{a}_1 - a_{0,1}\right)^2,\\
		\label{eq:mis_con_ctd_const_3}
		&\qquad\quad\qquad \text{SSE}_2=\sum_{\forall i\in\mathcal T}(1-z_{i})\left(y_i - \bm{m}_i^\intercal \bm{a}_2 - a_{0,2}\right)^2,\\
		\label{eq:mis_con_const_conti}
		&\qquad\quad\qquad \bm{a}_1 - \bm{a}_2 - \bm{w} = 0,
		\quad a_{0,1} - a_{0,2} - w_0 = 0,
	\end{align}
\end{subequations}
where $\bm{w}$ is a normal vector and $w_0$ constant off-set of the separation hyperplane, respectively, $\bm{e}$ is a vector of the slack variables, $\bm{z}$ is a vector of binary parameters that results from the data labeling procedure with $z_i=1$ if $\bm{m}_i \in \mathcal R_1$ and $z_i=0$ if $\bm{m}_i \in \mathcal R_2$, SSE is the sum of squared errors, $\alpha$ is a weighting parameter for normal vector of the separation plane and $\beta$ is a weighting parameter for vector of the slack variables.

The combination of the SVM-based classification of the data with the individual sensor training is represented by~\eqref{eq:mis_con_obj}--\eqref{eq:mis_con_ctd_const_3}. The resulting optimization problem is extended with constraints~\eqref{eq:mis_con_const_conti} which ensure the continuity at the switch between the two models. This is achieved by establishing the intersection of model surfaces to coincide with the determined switching hyperplane. To prove this claim, we first notice the following equivalence regarding the models intersection condition:
\begin{equation}
\bm{m}^\intercal \bm{a}_1 + a_{0,1} = \bm{m}^\intercal \bm{a}_2 + a_{0,2} \  \Leftrightarrow \ \bm{m}^\intercal (\bm{a}_1 - \bm{a}_2) + a_{0,1} - a_{0,2} = 0.
\end{equation}
The mutual intersection of the models and the switching hyperplane is established by the following condition:
\begin{align}
\bm{m}^\intercal (\bm{a}_1 - \bm{a}_2) + a_{0,1} - a_{0,2} = \bm{m}^\intercal \bm{w} + w_0 \ \Leftrightarrow \ \notag\\ \Leftrightarrow \ \bm{m}^\intercal \underbrace{(\bm{a}_1 - \bm{a}_2 - \bm{w})}_{=0} + \underbrace{a_{0,1} - a_{0,2} - w_0}_{=0} = 0,
\end{align}
from which~\eqref{eq:mis_con_const_conti} follows and which completes the proof.
Note that, we present the formulation of MIS$_\text{con}$ in the simplest form (two-model MIS, OLSR setup), for brevity, yet it is possible to extend easily this formulation to multiple models and other training approaches (see Section~\ref{sec:sis_train}).

As the a priori data labeling can be inappropriate for the design of a continuous MIS, we allow small violations of the labeling using the slack variables $\bm{e}$ in~\eqref{eq:mis_con_ctd_const_1}. We also consider that the user can aim at giving up some portion of model (training) accuracy for the better separation by widening the separation band. The latter feature is established by minimizing $\|\bm{w}\|_2^2$ in~\eqref{eq:mis_con_obj}. The described features can be enforced/weakened by tuning the positive weights $\alpha$ and $\beta$.

\paragraph{Illustrative example (continued)} The visualization of MIS$_\text{con}$ on the $PCT$ dataset can be seen in Figure~\ref{fig:pct_MIS_con_2cls}. Unlike the MIS$_\text{SotA}$ approach (Figure~\ref{fig:pct_MIS_SotA_2cls}), there is no discrepancy at the intercept between the designed models (yellow and dark green surfaces) of MIS$_\text{con}$. This confirms that the proposed approach ensures continuity when switching between designed models. The accuracy of MIS$_\text{con}$ (RMSE~($\mathcal T$)=0.079) is significantly decreased compared to MIS$_\text{SotA}$ (RMSE~($\mathcal T$)=0.032). This is a price to pay for MIS continuity and a design trade-off. 

Naturally, the continuity constraints~\eqref{eq:mis_con_const_conti} can be relaxed and introduced as soft constraints should one be willing to make the trade-off explicit for the MIS design. The accuracy and continuity of the MIS$_\text{con}$ model can then be effectively tuned by varying the weights $\alpha$ and $\beta$ according to the fidelity of a priori labeling and a desired level of discontinuity. The other way to improve the performance of the MIS$_\text{con}$ approach represents an implementation of the optimization of a priori labeling into the MIS design, which is further explored in the following text.

\subsection{Design of a Continuous MIS with Optimized Data Labeling}\label{sec:mis_con_lab}
In order to mitigate the inaccuracies caused by the a priori labeling of the training dataset, we propose the approach to design MIS with optimized data labeling (MIS$_\text{con,lab}$). This approach searches directly for the optimal data labeling by adding $\bm{z}$ among the optimized variables in~\eqref{eq:mis_con_obj}. The resulting optimization problem is following:
\begin{subequations}\label{eq:mis_con_lab}
	\begin{align}\label{eq:mis_con_lab_obj}
		&\min_{\substack{\bm{z}\in\{0,1\}^n, \bm{w}, w_0, \bm{e}\geq0\\\bm{a}_1, a_{0,1}, \bm{a}_2, a_{0,2}}}\text{SAE}_1 + \text{SAE}_2 + \alpha\|\bm{w}\|_1 + \beta\|\bm{e}\|_1\\
		\label{eq:mis_con_lab_const_1}
		&\qquad\text{s.t. }\quad (2z_{i}-1)\left(\bm{m}_i^\intercal \bm{w} + w_{0}\right) \geq 1 - e_{i}, \quad \forall i \in \{1, 2, \ldots, n\},\\
		\label{eq:mis_con_lab_const_2}
		&\qquad\quad\qquad \text{SAE}_1=\sum_{\forall i\in\mathcal T}z_{i}\left|y_i - \bm{m}_i^\intercal \bm{a}_1 - a_{0,1}\right|,\\
		\label{eq:mis_con_lab_const_3}
		&\qquad\quad\qquad \text{SAE}_2=\sum_{\forall i\in\mathcal T}(1-z_{i})\left|y_i - \bm{m}_i^\intercal \bm{a}_2 - a_{0,2}\right|,\\
		\label{eq:mis_con_lab_const_4}
		&\qquad\quad\qquad \bm{a}_1 - \bm{a}_2 - \bm{w} = 0,
		\quad a_{0,1} - a_{0,2} - w_0 = 0,
	\end{align}
\end{subequations}
where SAE is a sum of absolute errors.

Although a formulation similar to~\eqref{eq:mis_con} with SSE-based objective can be reused here, we adopt the SAE criterion to reduce the complexity. In a similar fashion, 2-norm is replaced for 1-norm to regularize the normal vector of the separating hyperplane. This is a standard approach~\citep{son_2002}. The optimization problem~\eqref{eq:mis_con_lab} can thus be transformed to a mixed-integer linear program (MILP). The transformation uses: (a) the epigraph reformulation~\citep{milano_2012} of the absolute value, (b) the big-M method~\citep{griva_2008} to linearize the bilinear constraints. As the variables $\bm{z}$ are binary, the big-M method does not require any new integer variables. If SSE was used in the objective function, the optimization problem would turn into mixed-integer nonlinear program (MINLP), which might be challenging especially when $n$ is high.

The problem~\eqref{eq:mis_con_lab} is primarily used to determine the data labels, which refer to how the training data is distributed and how the validity regions of the model are established. In principle, this approach parameterizes the choice of data distributions into model classes and thus makes the design independent of the used clustering algorithm. Once the values of $\bm{z}$ (data labels) are fixed, the final training of MIS models is performed by solving~\eqref{eq:mis_con} using the SSE criterion. This ensures a fair comparison with other SIS and MIS approaches.

This two-step approach does not require a priori labeling of the training set and can provide an optimal MIS at the expense of increased computational burden. The optimization problem for the MIS$_\text{con,lab}$ design increases by one binary optimized variable per training data point. Therefore, the proposed approach is limited to relatively small-scale problems (tens to hundreds of measurements). However, this is typically sufficient for the design of inferential sensors, where only a limited number of measurements is available for the desired (hard-to-measure) variable. Additionally, if a large dataset is available, a smaller size of the training dataset can be selected based on appropriate information criteria, similar to optimal design of experiments or sampling for surrogate model building~\citep{kamath_2022}.

\begin{figure}
	\centering
	\begin{subfigure}[b]{.48\textwidth}
		\includegraphics[width=\linewidth]{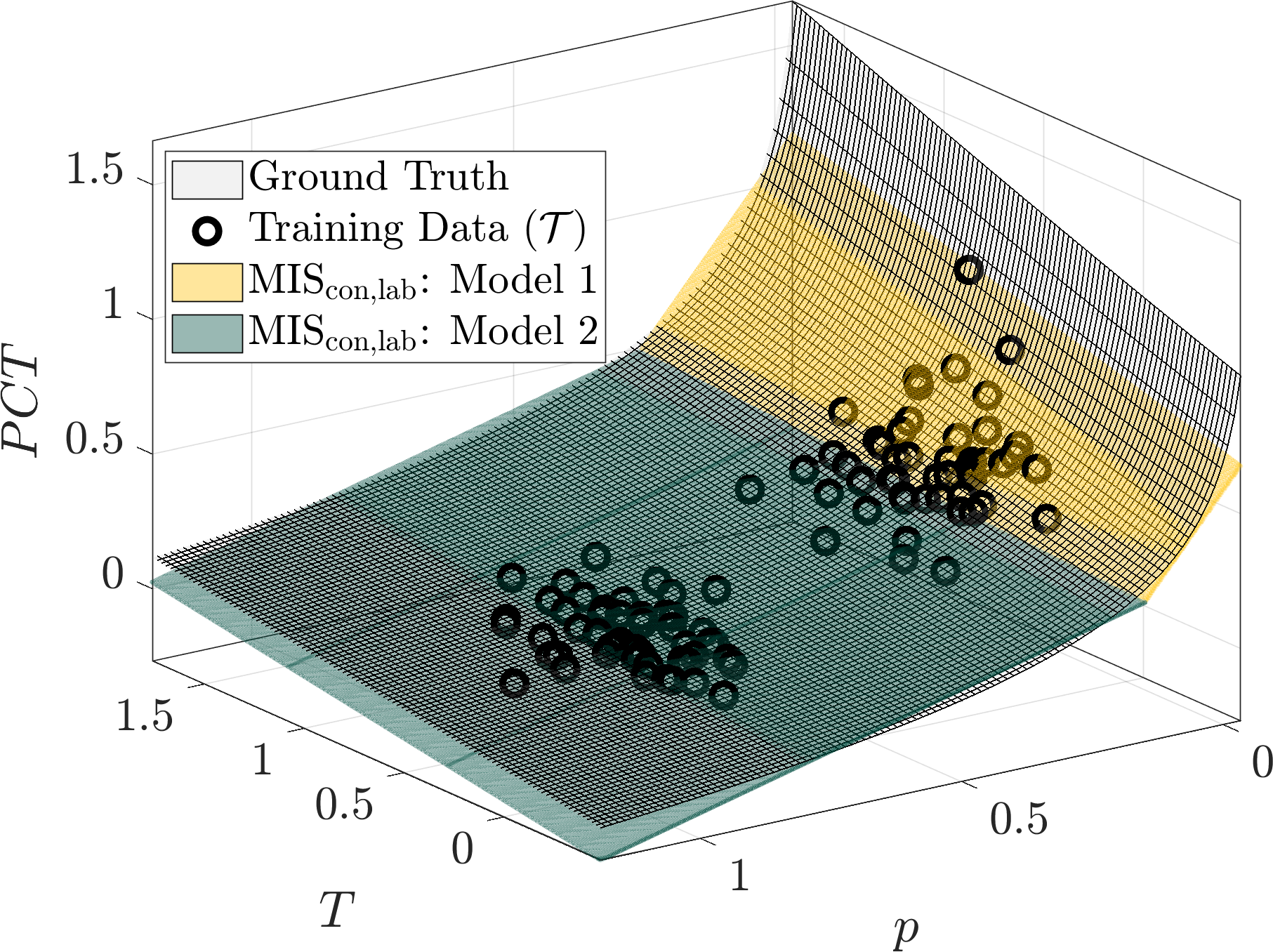}
		\caption{MIS$_\text{con,lab}$ (RMSE~$(\mathcal T)=0.023$) designed on the $PCT$ dataset with two distinct clusters.}
		\label{fig:pct_MIS_con_lab_2cls}
	\end{subfigure}\hfill
	\begin{subfigure}[b]{.48\textwidth}
		\includegraphics[width=\linewidth]{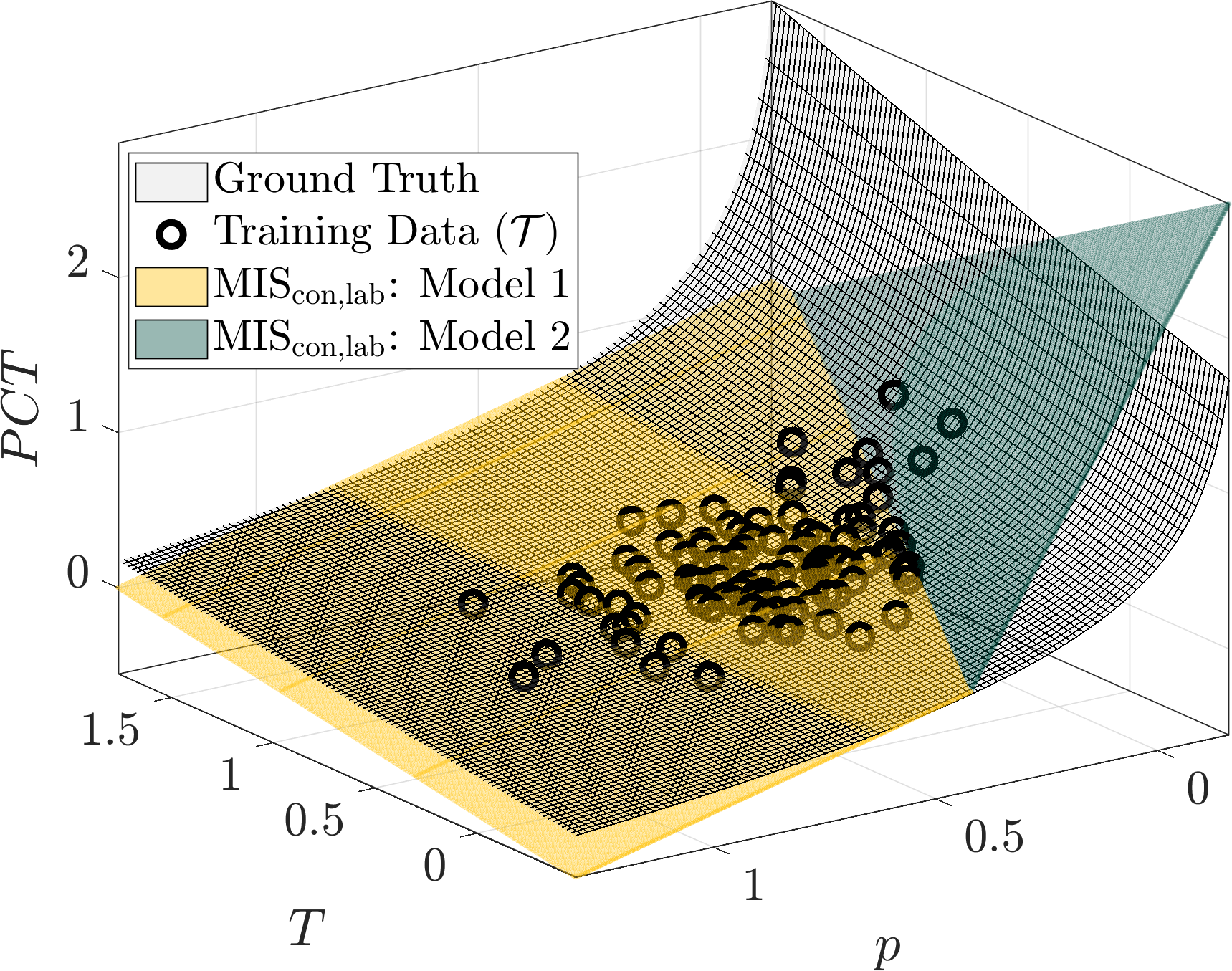}
		\caption{MIS$_\text{con,lab}$ (RMSE~$(\mathcal T)=0.043$) designed on the $PCT$ dataset with indistinguishable clusters.}
		\label{fig:pct_MIS_con_lab_rnd}
	\end{subfigure}
	\caption{The ground truth model of $PCT$ with MIS$_\text{con,lab}$ designed on different datasets.}
	\label{fig:pct_MIS_con_lab}\vspace{0cm}
\end{figure}

\paragraph{Illustrative example (continued)} Figure~\ref{fig:pct_MIS_con_lab} shows the design of MIS$_\text{con,lab}$ on the different $PCT$ datasets. The results in Figure~\ref{fig:pct_MIS_con_lab_2cls} show that the designed MIS$_\text{con,lab}$ has a high degree of flexibility and precision on the $PCT$ dataset with two distinct clusters. The accuracy of this sensor (RMSE~$(\mathcal T)=0.026$) outperforms MIS$_\text{SotA}$ (RMSE~$(\mathcal T)=0.032$) and MIS$_\text{con}$ (RMSE~$(\mathcal T)=0.079$) shown in Figure~\ref{fig:pct_MIS_SotA_vs_MIS_con_2cls}. The accuracy improvement of MIS$_\text{con,lab}$ is ensured by optimizing a priori labeling instead of using $k$-means clustering. This can be indicated by comparing classification in Figure~\ref{fig:pct_MIS_con_lab_2cls} against Figure~\ref{fig:pct_MIS_SotA_vs_MIS_con_2cls}. The results from Figure~\ref{fig:pct_MIS_con_lab_rnd} indicate that the sensor returned by the MIS$_\text{con,lab}$ approach is designed effectively even when the considered dataset has no distinguishable clusters. This is typical for industrial datasets due to the presence of the significant level of noise and multitude of similar operating points.

\section{Case Studies}
\label{sec:case_studies}
The design of single-model inferential sensors (SIS) and multi-model inferential sensors (MIS) is elaborated on two case studies. Both case studies have an industrial character and practical relevance. The first case study features a pressure compensated temperature $PCT$, which is briefly explored in Section~\ref{sec:sis_train}. The purpose of this case study is to analyze the impact of data quality on the MIS design in multitude of simulations. The second case study involves an industrial dataset from the VGH unit, which is a part of the oil refinery Slovnaft, a.s. in Bratislava, Slovakia. This case study validates the applicability of the proposed MIS design approaches in practice.

\subsection{Implementation Details}\label{sec:impl_details}
The presented design methods are implemented in MATLAB R2022a. To solve the involved optimization problems, we use the Yalmip package~\citep{lofberg_2004} and Gurobi solver~\citep{gurobi_2020}. All the numerical results and graphical representations consider the normalization of variables within the interval $[0,1]$ in both case studies. The normalization (scaling) parameters are not disclosed for the dataset from the VGH unit due to data confidentiality. For comparison, we also use a single-model sensor represented via neural network. The neural network is trained using Matlab's Neural Network toolbox.

Prior to the inferential sensor design, the entire available dataset is divided into training and testing (unseen) datasets. The information contained within the training and testing datasets is one of the decisive factors directly affecting the performance of the designed inferential sensors. Therefore, the effect of various ways of dividing the data into training and testing datasets on the SIS and MIS performances is further investigated on the $PCT$ dataset in Section~\ref{sec:ss_design_for_pct}. The inferential sensor design on the industrial dataset from the VGH unit is based on the random distribution of measurements in the training and testing datasets.

The design of MIS$_\text{SotA}$ performs the a priori labeling by using $k$-means clustering. Subsequently, the linear classifier of MIS$_\text{SotA}$ is designed by SVM. Finally, the model parameters within MIS$_\text{SotA}$ models are calculated by OLSR. The MIS$_\text{con}$ design is performed according to~\eqref{eq:mis_con}. In order to reduce the computational effort, MIS$_\text{con}$ is initialized by the results from~\eqref{eq:mis_con} considering SAE instead of SSE within the objective function~\eqref{eq:mis_con_obj}. Subsequently, the MIS$_\text{con,lab}$ design from~\eqref{eq:mis_con_lab} is initialized by the results from MIS$_\text{con}$.

\subsection{Design of Inferential Sensors for Pressure-Compensated Temperature}\label{sec:ss_design_for_pct}
We use the datasets generated by simulating the nonlinear model of $PCT$ represented by~\eqref{eq:pct_model} with respect to the parameters and specifications introduced in Section~\ref{sec:sis_train}. We use this ideal case study to examine the impact of various factors on the performance of the inferential sensor. Specifically, in this section, we analyze the impact of two factors on the SIS and MIS designs: (1) the method of data distribution into the training and testing datasets and (2) the noise variance in the output variable. Overall, the studied $PCT$ dataset involves 620 measurements, which are equally distributed between the training and testing datasets. We use this setup for comparison purpose as it will be used in the second case study due to nature of the considered plant operation.

The considered case study involves two input variables ($p$ and $T$) and one output variable ($PCT$), and therefore, it is unnecessary to consider the SIS approaches with advanced input structure selection (i.e., SIS$_\text{PCR}$, SIS$_\text{PLSR}$, SIS$_\text{LAS}$, and SIS$_\text{SS}$). We use a soft-sensor designed by a neural network (designated as SIS$_\text{NN}$) to compare the linear and multi-linear sensors against a nonlinear SIS. The structure of the network was optimized using cross-validation. The best structure involves fully-connected neural network with two inputs, no hidden layer, and one neuron in the output layer with hyperbolic tangent as an activation function. The set of compared inferential sensors in this section involves SIS$_\text{OLSR}$, MIS$_\text{SotA}$, MIS$_\text{con}$, and  MIS$_\text{con,lab}$.

\begin{figure}
	\centering
	\begin{subfigure}[b]{.495\textwidth}
		\includegraphics[width=\linewidth]{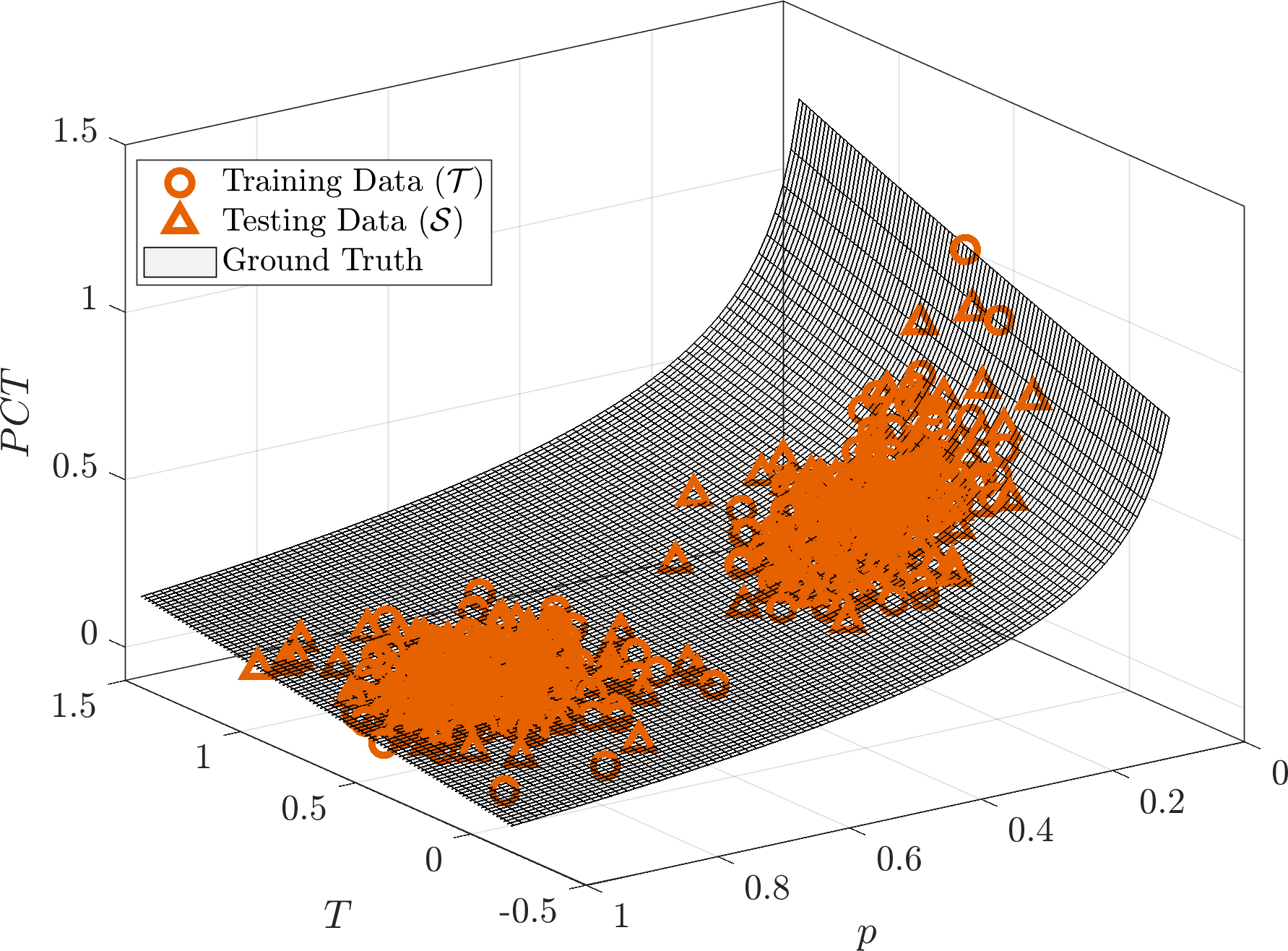}
		\caption{The desirable scenario.}
		\label{fig:pct_TR_TS_des}
	\end{subfigure}
	\begin{subfigure}[b]{.495\textwidth}
		\includegraphics[width=\linewidth]{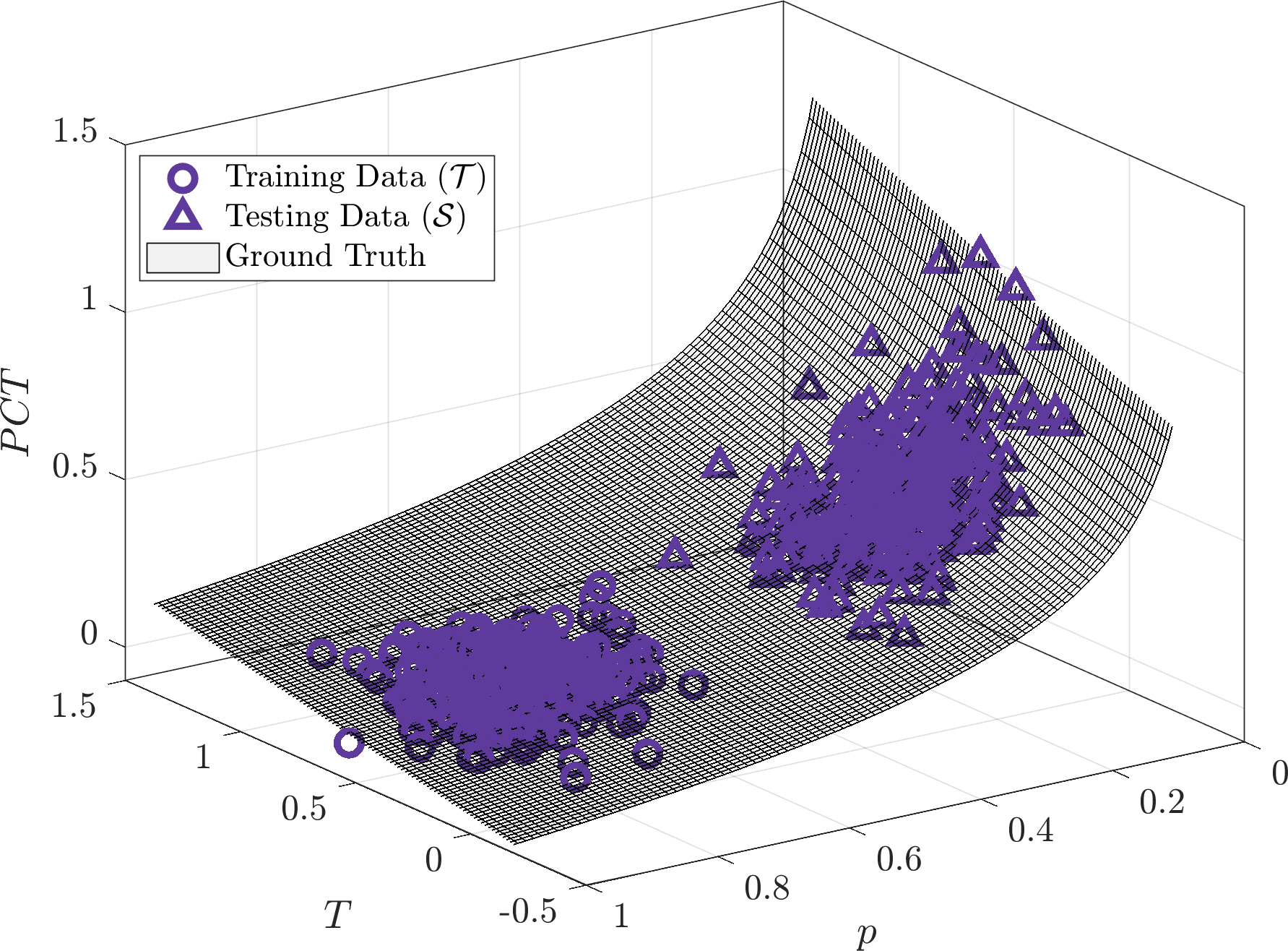}
		\caption{The undesirable scenario.}
		\label{fig:pct_TR_TS_undes}
	\end{subfigure}
	\caption{The comparison of studied distributions of the $PCT$ data into training and testing datasets.}
	\label{fig:pct_TR_TS}
\end{figure}

To analyze the impact of the data distribution into the training and testing datasets, we generate datasets from the $PCT$ model according to two different scenarios. The first scenario, or desirable scenario, considers that the $PCT$ is sampled in two different operating regimes and measurements from both operating regimes are available in the training dataset, as illustrated in Figure~\ref{fig:pct_TR_TS_des}. This scenario occurs relatively frequently in the industry. It assumes that the process operates only within known operating regimes, which is desired for the inferential sensor design. The second scenario, or undesirable scenario, assumes the same operating regimes as the first scenario, but the training dataset involves measurements from one operating regime only, and the testing dataset involves measurements from the other operating regime, as shown in Figure~\ref{fig:pct_TR_TS_des}. This scenario represents an undesirable, yet not unlikely, situation in the industry, when the process operates within a new operating state after the inferential sensor design. 

\begin{figure}
	\centering
	\includegraphics[width=\linewidth]{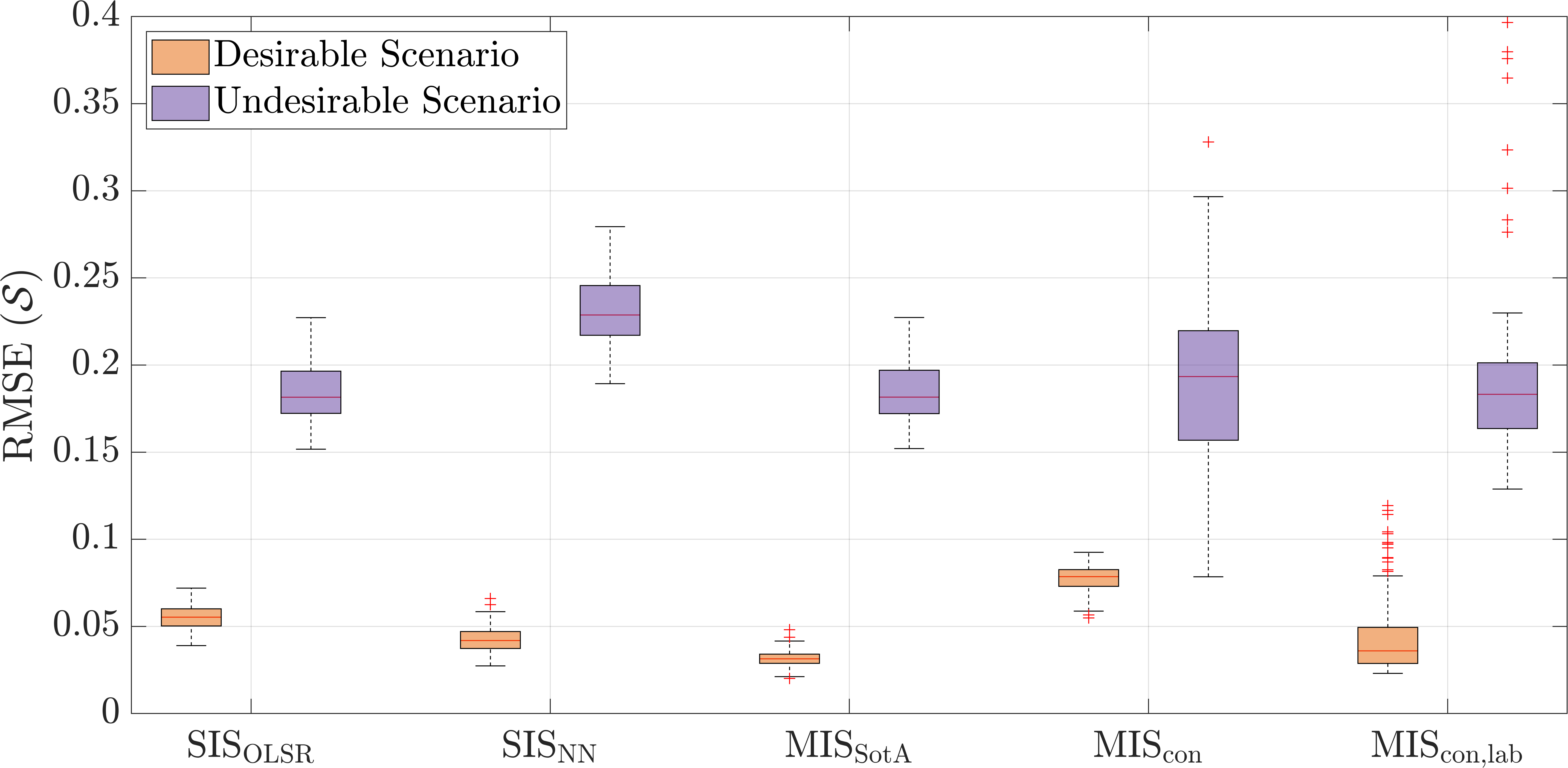}
	\caption{The statistical comparison of RMSE~($\mathcal S$) of designed inferential sensors (SIS$_\text{OLSR}$, SIS$_\text{NN}$, MIS$_\text{SotA}$, MIS$_\text{con}$ and MIS$_\text{con,lab}$) involving 100 different datasets for each studied scenario.}
	\label{fig:pct_TR_TS_comp}
\end{figure}

The comparison in Figure~\ref{fig:pct_TR_TS_comp} shows a statistical evaluation of the results obtained over 100 different datasets for each studied scenario. The datasets consider two classes of measurements, as shown in Figure~\ref{fig:pct_TR_TS}, with different random distributions of measurements. The noise considered within the output variable is a random variable from $\mathcal N(\mu_\text{noise},\sigma_\text{noise}^2) = \mathcal N(0,5^2)$, where the value of the standard deviation $\sigma_\text{noise}$ represents 0.67\,\% ($\sigma_\text{noise}/\mu_{PCT} \cdot 100\,\%$) of an averaged value from the original (prior to the normalization) output variable. The boxes in Figure~\ref{fig:pct_TR_TS_comp} represent the 25$^\text{th}$ and 75$^\text{th}$ percentiles of the RMSE reached on the testing test. The red line within the box represents the median value of the considered set of results. The red crosses represent the statistical outliers. For each designed inferential sensor (i.e., SIS$_\text{OLSR}$, SIS$_\text{NN}$, MIS$_\text{SotA}$, MIS$_\text{con}$, and MIS$_\text{con,lab}$), there is a pair of orange (desirable scenario) and violet (undesirable scenario) boxes in Figure~\ref{fig:pct_TR_TS_comp}. The trained MIS models involve two models because of the nature of the dataset (in the desirable scenario).

The results presented in Figure~\ref{fig:pct_TR_TS_comp} demonstrate that the designed inferential sensors exhibit better performance on datasets from the desirable scenario (represented by the orange boxes) compared to those from the undesirable scenario (represented by the violet boxes). These findings support our initial assumption regarding the impact of these scenarios on the performance of inferential sensors.

\subsubsection{Analysis of the Desirable Scenario}
The results from the desirable scenario further indicate a high degree of robustness (or low variance) in the SIS$_\text{OLSR}$ and SIS$_\text{NN}$ accuracy, as seen by the small height of the corresponding orange boxes in Figure~\ref{fig:pct_TR_TS_comp}. On the other hand, the accuracy of SIS$_\text{OLSR}$ is significantly outperformed by MIS$_\text{NN}$, which results from the ability of neural network to represent nonlinear transformation from input to output. The accuracy of SIS$_\text{OLSR}$ is also significantly outperformed by MIS$_\text{SotA}$. The performance variance of MIS$_\text{SotA}$ is even smaller compared to that of SIS$_\text{OLS}$ and SIS$_\text{NN}$. MIS$_\text{SotA}$ even provides better performance than SS$_\text{NN}$, which can be attributed to the number of training data, where neural network cannot identify the nonlinearity in the output to a sufficient extent. The outstanding performance of MIS$_\text{SotA}$ can be attributed to the nature of the datasets in the desirable scenario (as shown in Figure~\ref{fig:pct_TR_TS_des}) with distinguishable classes. One class involves measurements that precisely explain the behavior of $PCT$ in the (almost) linear section, while the other class involves mostly measurements from the highly nonlinear section of the $PCT$ model range. Therefore, a priori labeling within the MIS$_\text{SotA}$ approach ($k$-means clustering) provides appropriate data labels for the subsequent design of inferential sensors. One can conclude that multi-model linear sensor would be a reasonable choice in this case study.

The results of MIS$_\text{con}$ (as shown in the corresponding orange box in Figure~\ref{fig:pct_TR_TS_comp}) indicate a small variance in accuracy comparable to that of SIS$_\text{OLSR}$ and SIS$_\text{NN}$. However, MIS$_\text{con}$ achieves the lowest accuracy compared to other designed inferential sensors on the datasets from the desirable scenario. This suggest that the a priori labeling is not optimal \wrt design of a continuous multi-model sensor. Further analysis of the results reveals that the MIS$_\text{con}$ accuracy (considering the desirable scenario) can be significantly improved by the MIS$_\text{con,lab}$. The results in the desirable scenario exhibit higher variability in the performance compared to all other designed inferential sensors. The main reason for this is the complexity of the optimization problem that needs to be solved, which can lead to numerical inaccuracies. This assumption is supported by the increased occurrence of outliers in the MIS$_\text{con,lab}$ results in Figure~\ref{fig:pct_TR_TS_comp} (represented by the red crosses). Despite the increased variance of the accuracy, MIS$_\text{con,lab}$ achieves comparable accuracy to MIS$_\text{SotA}$ in the majority of the cases. This is notable given the similar values of the median (represented by the red lines) of these MISs.
 
\subsubsection{Analysis for the Undesirable Scenario}
The results from the undesirable scenario (violet boxes) indicate that SIS$_\text{OLSR}$ and MIS$_\text{SotA}$ show similar performance and are the most accurate among the studied sensors. Unlike the observations from the desirable scenario, the accuracy of MIS$_\text{SotA}$ is only slightly higher than that of SIS$_\text{OLSR}$. On the other hand, SIS$_\text{NN}$ exhibits poor extrapolation properties and performs the worst among all studied sensors. Despite the challenge of making extrapolated predictions, MIS$_\text{SotA}$ still performs the best and exhibits the lowest variance of accuracy among all the sensors.

We can observe very high variability in the MIS$_\text{con}$ accuracy. This stems from the requirement to design continuous models, which restricts the rotation and angle between the models. The results suggest that the variance and accuracy of MIS$_\text{con}$ can be improved by using the MIS$_\text{con,lab}$ approach. However, the possibilities of MIS$_\text{con,lab}$ are limited due to the nature of the training/testing dataset. As a result, we observe similar accuracy of MIS$_\text{con,lab}$ compared to SIS$_\text{OLSR}$ and MIS$_\text{SotA}$, as indicated by the median values. We also note the occurrence of low-accuracy outliers within the MIS$_\text{con,lab}$ results, similar (yet more pronounced) to the desirable scenario.

This scenario represents an extreme case where the estimated behavior is completely unknown during the training phase. However, in real-world applications, it is more likely that the studied problem will fall in-between the desirable and undesirable scenarios. Therefore, we can conclude that the results in Figure~\ref{fig:pct_TR_TS_comp} confirm that MIS$_\text{SotA}$ and MIS$_\text{con,lab}$ have the potential to more accurately capture nonlinear behavior compared to the linear SIS$_\text{OLS}$, as observed in Figures~\ref{fig:pct_MIS_SotA_2cls} and~\ref{fig:pct_MIS_con_lab_2cls}. The difference in accuracy between linear MIS and SIS would be even more pronounced in more nonlinear case studies.

\subsubsection{Analysis of the Measurement Noise}
The previous analysis shows that the random distribution of data into the training and testing datasets (the desirable scenario) should provide sufficient informative content for the training dataset. Therefore, we use this distribution in the following analysis focused on the impact of noise (in the output variable) on the performance of the designed inferential sensors. The set of noise variances is selected as $\sigma_\text{noise} \in \{0.1,\,2.5,\,5,\,10,\,25,\,50\}$. The minimum, respectively maximum, considered noise variance ($\sigma_\text{noise}=0.1$, respectively $\sigma_\text{noise}=50$) represent approximately 0.1\,\%, respectively 6.7\,\%, of the mean value of the output variable from the considered datasets. To provide more representative results, we present the median RMSE~($\mathcal S$) value from 100 different realizations of noise for each studied $\sigma_\text{noise}$. The purpose of this analysis is to investigate the practical applicability of the studied MISs concerning data quality, specifically focusing on the influence of noise. This aspect becomes crucial in industrial applications, where the accuracy of measurements is closely linked to the quality of available measurement and sensing devices.

\begin{figure}
	\centering
	\includegraphics[width=\linewidth]{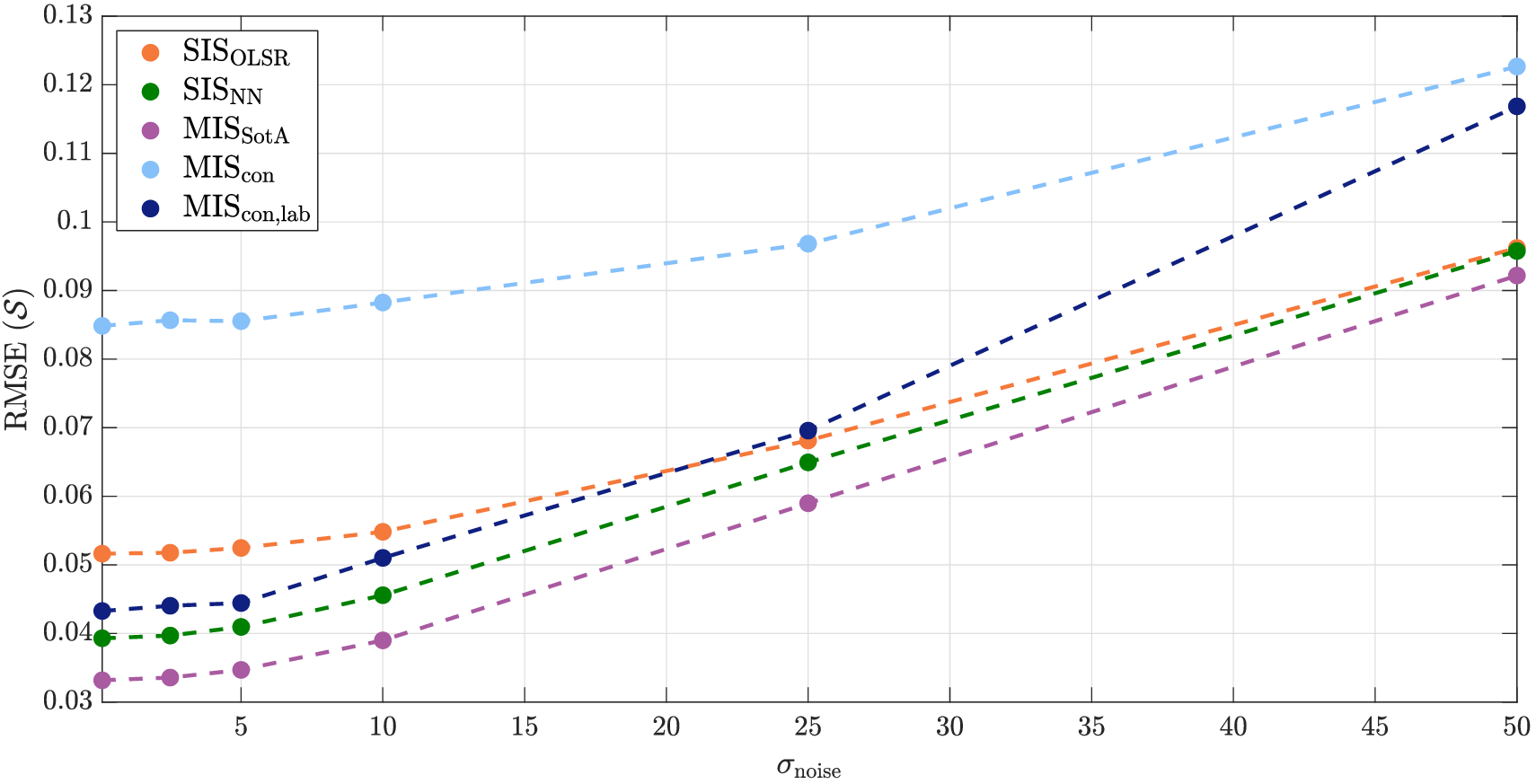}
	\caption{The averaged RMSE~($\mathcal S$) value for designed inferential sensors (SIS$_\text{OLSR}$, MIS$_\text{SotA}$, MIS$_\text{con}$, and MIS$_\text{con,lab}$) from 100 different realizations of the noise for each studied $\sigma_\text{noise}$ within the output variable.}
	\label{fig:pct_noise_comp}
\end{figure}

The impact of the noise variance (within the output variable) on the accuracy (RMSE~($\mathcal S$)) of the designed inferential sensors is illustrated in Figure~\ref{fig:pct_noise_comp}. In general, an increase in $\sigma_\text{noise}$ leads expectedly to an increase in RMSE~($\mathcal S$). The performance of SIS$_\text{OLSR}$ (orange points) appears to be relatively robust as the overall RMSE increase is the smallest among all the studied sensors iver the entire studied noise magnitude interval. Regarding SIS$_\text{NN}$ and MIS$_\text{SotA}$ (green and magenta points, respectively), the previously indicated trend is confirmed. The superior performance of MIS$_\text{SotA}$ is expected due to the nature of the considered datasets (see Figure~\ref{fig:pct_TR_TS_des}) and nonlinearity of the ground-truth model, as discussed above. The performance of SIS$_\text{NN}$ sees a deterioration in performance with increasing noise level and even reaches the performance of a linear sensor (SIS$_\text{OLSR}$) for the highest value.

The (low) accuracy of MIS$_\text{con}$ (represented by pale blue points) follows the trends already observed. It is interesting to see that increased noise levels do not have a dramatic impact on performance, so the previously observed large performance variance can be mitigated by a considerate training, involving shuffling of training data and cross-validation. The results of MIS$_\text{con,lab}$ (blue points) indicate that the accuracy of MIS$_\text{con}$ can be significantly improved by optimizing the data labeling. MIS$_\text{con,lab}$ outperforms SIS$_\text{OLSR}$ when $\sigma_\text{noise}<20$. The accuracy of MIS$_\text{con,lab}$ decreases steeper compared to other approaches when $\sigma_\text{noise}>20$. The increased flexibility of MIS$_\text{con,lab}$ leads to an increased tendency for this approach to explain the noise within the training dataset, especially for significant noise variances. In other words, the MIS$_\text{con,lab}$ approach has an increased tendency for overfitting for large noise levels.

\subsection{Design of Inferential Sensors for Vacuum Gasoil Hydrogenation Unit}\label{sec:ss_design_for_vgh}
\begin{figure}
	\centering
	\begin{tikzpicture}
		\node[anchor=south west,inner sep=0] (image) at (0,0) {\includegraphics[width=\linewidth]{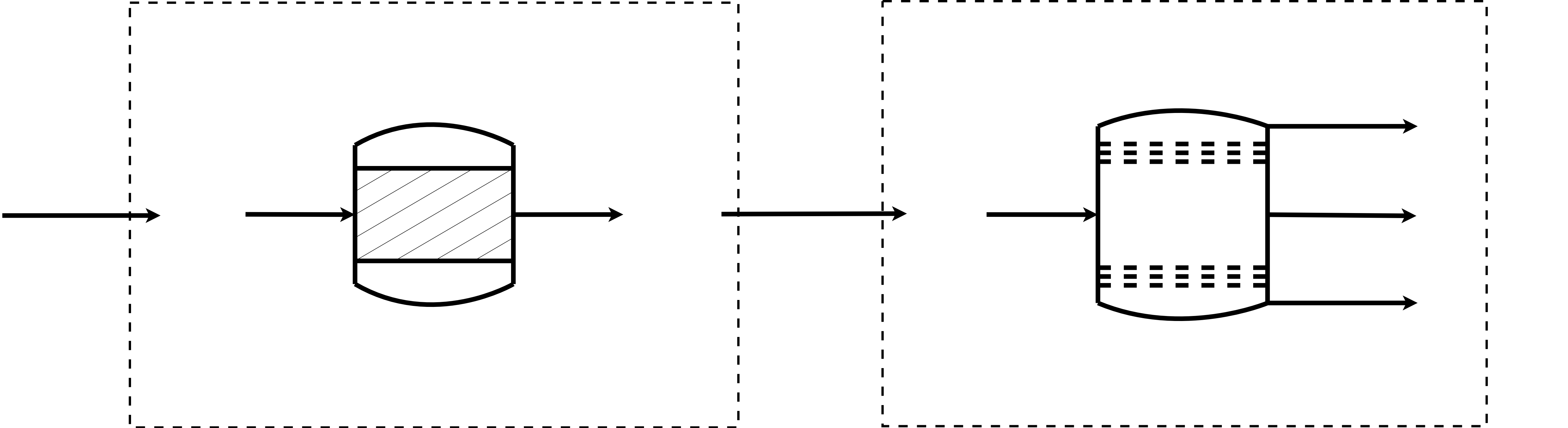}};
		\node at (78pt,97pt) {\emph{Reaction Section}};
		\node at (280pt,97pt) {\emph{Fractionation Section}};
		\node at (16pt,59pt) {Feed};
		\node at (110pt,22pt) {Main Reactor};
		\node at (328pt,81pt) {GF};
		\node at (333pt,59pt) {HGO};
		\node at (333pt,37pt) {other};
		\node at (300pt,18pt) {\parbox{5cm}{\centering Product Fractionator}};
	\end{tikzpicture}
	\caption{A schematic diagram of the VGH unit.}
	\label{fig:VGH_plant_scheme}
\end{figure}
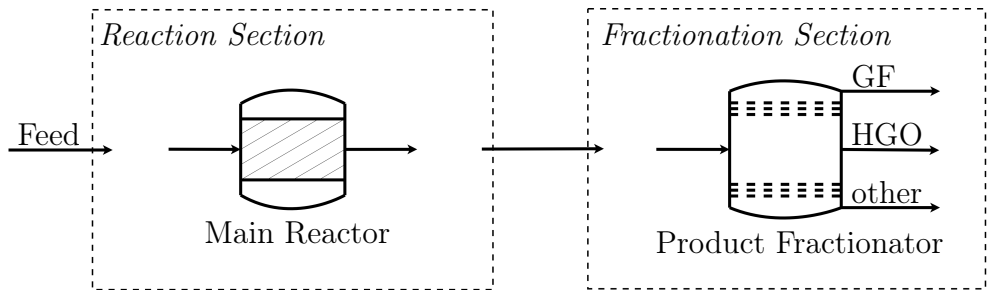

The Vacuum Gasoil Hydrogenation (VGH) unit is an essential part of the oil refinery Slovnaft, a.s. in Bratislava, Slovakia. This unit (schematic in Figure~\ref{fig:VGH_plant_scheme} processes the vacuum distillates in two consecutive sections. The first section is a high-pressure reaction that significantly reduces the concentration of the impurities (e.g., nitrogen and sulfur) in the feed mixture. The second section is a low-pressure fractionation, where a product fractionator (main distillation column) separates the pre-treated vacuum distillates into a gasoline fraction (GF), a hydrogenated gasoil (HGO), and other products.

The available industrial dataset involves measurements for 24 months of the VGH unit operation. The output (desired) variable to be estimated by inferential sensors is the purity of the HGO product, represented by 95\,\% point of the distillation curve $T_{95\,\%;\text{HGO}}$. The lab analysis of the HGO product is executed approximately once per day, and therefore, there are 621 measurements of the output variable available for the inferential sensor design. The input variables are measured every minute by online sensors. In order to reduce the impact of the measurement noise, the minutely measurements are replaced by the averaged measurements from 30-minute intervals. The resulting input dataset involves 27,324 measurements. The available output measurements are collocated time-wise within the matching 30-minute interval among the available input data.

The set of input candidates involves the following 35 variables:
\begin{equation}\label{eq:vgh_inputs}
	\begin{aligned}
		\bm{m}_i = \left(\right. & T_\text{F,5p}, T_\text{F,50p}, T_\text{F,95p}, T_\text{col}, T_\text{vap,t}, T_\text{ex,1}, T_\text{ex,2}, \\ 
		& T_\text{ex,3}, T_\text{ex,4}, T_\text{wb,1}, T_\text{wb,2}, T_\text{wb,3}, T_\text{wb,4}, T_\text{wb,5},\\
		& x_{\text{F,N}_2}, x_\text{F,S}, x_{\text{H}_2}, \text{rt}_1, \text{rt}_2, \text{rt}_3, \text{rt}_4, \text{rt}_5, \text{rt}_6, \text{rt}_7, \\
		& F_\text{F}, F_\text{F,rec}, F_\text{R}, F_\text{h}, p_\text{col}, L_\text{reb}, \text{vo}_\text{R}, \text{vo}_\text{h,1}, \text{vo}_\text{h,2},\\
		& PCT_\text{HGO}, \left. PCT_\text{GF} \right)_i^\intercal,
	\end{aligned}
\end{equation}
where $T_\text{F,5p/50p/95p}$ is a temperature from the feed stream distillation curve for various points (e.g., 5\,\%, 50\,\%, 95\,\%) of the distillation curve, $T_\text{col}$ is a temperature within the column section, $T_\text{vap,t}$ is a temperature of the vapor phase on the top of the column, $x_{\text{F,N}_2/\text{S}}$ is a content of the nitrogen/sulfur in the feed stream, $F_\text{F/R}$ is a flowrate of the feed/reflux stream, $F_\text{F,rec}$ is reconciled flowrate of the feed stream, $F_\text{h}$ is a flowrate of heat medium to the reboiler, $p_\text{col}$ is a pressure within the column section, $L_\text{reb}$ is a liquid level in the reboiler, vo$_\text{R}$ is a valve opening of the reflux stream, vo$_\text{h,1--2}$ is a valve opening of the heating medium for the reboiler (1 -- input, 2 -- output) and $PCT_\text{HGO/GF}$ is a pressure-compensated temperature for HGO/GF stream from the product fractionator. Note that $PCT_\text{HGO/GF}$ are calculated using (undisclosed) industrial data for the sake of fair comparison with plant current standards. The variables described above are measured directly at or very close to the product fractionator (low-pressure fractionation section). The remaining variables in~\eqref{eq:vgh_inputs} are located in the high-pressure reaction section, where $T_\text{ex,1--4}$ is a temperature at the various locations of the catalytic reactors monitoring (exothermic) reaction, $T_\text{wb,1--5}$ is weighted average bed temperature at the different locations of the catalytic reactors, $x_{\text{H}_2}$ represents the content of the hydrogen in the reaction section, rt$_{\text{1--7}}$ is a ratio of the gas/liquid phases in the different areas of the reaction section.

The supervised learning methods used for inferential sensor design require a paired input-output dataset, where the inputs and outputs correspond to the same measurement time. In this case, the input and output datasets consist of 621 measurements, which are subsequently divided randomly into training (311 measurements) and testing (310 measurements) sets. The equal division between the training and testing datasets stems from the nature of the plant operation that covers a large operating window. We use this setup to guarantee a fair comparison and representativeness of the testing data. It is worth noting that the dataset is comparable in size to the $PCT$ dataset (620 measurements) used in the previous case study (Section~\ref{sec:ss_design_for_pct}). This allows us to explore any similarities between the results and conclusions of the case studies.

Currently, the refinery has implemented a univariate SIS ($n_\text{p}^*=1$) with the following structure:
\begin{equation}\label{eq:vgh_SIS_ref}
	\hat{y}_i = PCT_{\text{HGO},i} \cdot a_1 + a_0.
\end{equation}
This structure is based on the expert knowledge of the operators and engineers in the refinery and is used as a reference in the comparison of designed inferential sensors in this study. Therefore, we refer to this sensor as SIS$_\text{Ref}$. The OLSR approach is used to evaluate the parameters in \eqref{eq:vgh_SIS_ref}.

The set of approaches considered for SIS design (see Section~\ref{sec:sis_train}) includes OLSR (SIS$_\text{OLSR}$), PCR (SIS$_\text{PCR}$), PLSR (SIS$_\text{PLSR}$), and LASSO (SIS$_\text{LAS}$). These approaches consider the entire set of input candidates ($n_\text{p}=35$) and search for the optimal input structure ($n_\text{p}^*$) based on their specific objectives. Similar to the previous case study, we use a neural network (designated as SIS$_\text{NN}$) to compare the linear and multi-linear sensors against a nonlinear SIS. The structure of the network was optimized using cross-validation. The same structure as in the previous case study (fully-connected neural network with 35 inputs, no hidden layer, and one neuron in the output layer with hyperbolic tangent as an activation function) was concluded to be the best.

The set of compared approaches for MIS design consists of MIS$_\text{SotA}$, MIS$_\text{con}$, and MIS$_\text{con,lab}$. The MIS approaches consider two candidate input structures: (1) the reference input structure (denoted as Ref) given by~\eqref{eq:vgh_SIS_ref} with $n_\text{p}^*=1$ ($PCT_\text{HGO}$) and (2) the input structure determined by the subset selection method with $n_\text{p}^*=2$ (denoted as SS) with $n_\text{p}^*=2$ ($PCT_\text{HGO}$ and $\text{vo}_\text{h,2}$). The SS input structure is determined using the MIS$_\text{con}$ approach for structures selected by the SS method with varying number of non-zero model parameters ($\tilde n_\text{p}$). We selected $n_\text{p}^*=2$ as a compromise between simplicity and performance in the RMSE~($\mathcal T$) criterion. The choice of simple structure has two merits. Firstly, expert knowledge from the refinery suggests to consider simple input structures (see reference structure in~\eqref{eq:vgh_SIS_ref}). Secondly, the MIS$_\text{con,lab}$ approach tends to overfit, as indicated in Section~\ref{sec:ss_design_for_pct}. To provide a fair comparison, the SIS set of approaches is extended by SIS$_\text{Ref}$ and SIS$_\text{SS}$, considering the same input structures as the MIS approaches. The MIS approaches involve two models as a compromise between simplicity and rate of overfitting. The $k$-means method suggested the number of clusters to be 2--5.

\begin{table}
	\caption{Comparison of the resulting number of input variables $n_\text{p}^*$ / principal components $n_\text{pc}^*$ (for PCR and PLSR) and accuracy (RMSE) of designed SISs on the training~($\mathcal T$) and testing~($\mathcal S$) industrial datasets.}
	\label{tab:vgh_SIS_res}\centering
	\begin{tabular}{c|c c c c c c c}
		\toprule
		                                    & SIS$_\text{OLSR}$ & SIS$_\text{PCR}$ & SIS$_\text{LAS}$ & SIS$_\text{Ref}$ & SIS$_\text{SS}$ & SIS$_\text{NN}$ \\ \midrule
		$n_\text{p}^*$/$n_\text{pc}^*$      & 31/-  & 35/6  & 13/-  & 1/-   & 2/-   & 35/- \\
		RMSE~($\mathcal T$)                 & 0.077 & 0.109 & 0.083 & 0.111 & 0.104 & 0.077\\
		RMSE~($\mathcal S$)                 & 0.145 & 0.105 & 0.097 & 0.1   & 0.095 & 0.105\\
		\bottomrule
	\end{tabular}
\end{table}

The resulting performance criteria of the studied single-model inferential sensors designed on the VGH dataset are shown in Table~\ref{tab:vgh_SIS_res}. The complexity of the inferential sensors is represented by the resulting number of input variables $n_\text{p}^*$ and the number of principal components $n_\text{pc}^*$ (for SIS$_\text{PCR}$).

The results indicate that SIS$_\text{OLSR}$ achieves the lowest RMSE~($\mathcal T$) value among all studied SISs, but it produces a relatively high RMSE~($\mathcal S$) value and a complex input structure ($n_\text{p}^*=31$), which suggests overfitting. The variance-covariance approach, SIS$_\text{PCR}$, indicates also high input structure complexity ($n_\text{p}^*=35, n_\text{pc}^*=6$). Performance of SIS$_\text{PLSR}$ is not included as it is practically the same as the performance of SIS$_\text{LAS}$. SIS$_\text{LAS}$ shows improved accuracy compared to the aforementioned two sensors and its input structure is less complex ($n_\text{p}^*=13$). SIS$_\text{Ref}$ ($n_\text{p}^*=1$) and SIS$_\text{SS}$ ($n_\text{p}^*=2$) have significantly lower complexity compared to other designed SISs. Based on these results, it is possible to improve the performance of SIS$_\text{Ref}$ (currently implemented in the refinery) by about 5\,\% using SIS$_\text{SS}$. This would only require to maintain one extra online sensor (compared to SIS$_\text{Ref}$), to ensure the accuracy and reliability of SIS$_\text{SS}$. Performance of SIS$_\text{NN}$ is at the lower end being similar to SIS$_\text{PCR}$. This is attributed to the lack of training data and to high level of noise, as indicated in the analysis done for the previous case study.

\begin{table}
	\caption{Comparison of the resulting number of input variables $n_\text{p}^*$ and accuracy (RMSE) of designed MISs on the training~($\mathcal T$) and testing~($\mathcal S$) industrial datasets.}
	\label{tab:vgh_MIS_res}\centering
	\begin{tabular}{c|c c c c c c}
		\toprule
		& \multicolumn{2}{c}{MIS$_\text{SotA}$} & \multicolumn{2}{c}{MIS$_\text{con}$} & \multicolumn{2}{c}{MIS$_\text{con,lab}$} \\ \midrule
		$n_\text{p}^*$                      & 1 (Ref) & 2 (SS) & 1 (Ref) & 2 (SS) & 1 (Ref) & 2 (SS) \\
		RMSE~($\mathcal T$)                 & 0.098   & 0.097  & 0.109   & 0.130   & 0.108   & 0.1   \\
		RMSE~($\mathcal S$)                 & 0.094   & 0.092  & 0.113   & 0.133  & 0.105   & 0.098  \\
		\bottomrule
	\end{tabular}
\end{table}

Table~\ref{tab:vgh_MIS_res} presents a comparison of the performance of MISs (MIS$_\text{SotA}$, MIS$_\text{con}$, and MIS$_\text{con,lab}$) designed for the VGH dataset. Each of these approaches shows the resulting quality of the designed inferential sensors using both input structures (Ref and SS), and their accuracy is evaluated according to the same criteria as in the case of SISs (Table~\ref{tab:vgh_SIS_res}). This enables a direct comparison of the results from the studied SIS and MIS approaches.

Table~\ref{tab:vgh_MIS_res} suggests that MIS$_\text{SotA}$ achieves the highest accuracy (RMSE~($\mathcal S$)) compared to other designed MISs and SISs, taking into account both input structures. These results confirm the excellent accuracy of the MIS$_\text{SotA}$ approach on the $PCT$ datasets (Section~\ref{sec:ss_design_for_pct}). Moreover, the accuracy of MIS$_\text{SotA}$ is higher with the SS input structure than with the Ref input structure, suggesting that the SS input structure is more appropriate for the MIS design. The MIS with the Ref input outperforms the currently implemented inferential sensor in the refinery (see SIS$_\text{Ref}$ in Table~\ref{tab:vgh_SIS_res}) by about 6\,\% and with the SS input structure by about 8\,\%.

The accuracy of MIS$_\text{con}$ appears worse compared to MIS$_\text{SotA}$ for both input structures. This aligns with observations made on the $PCT$ datasets (Section~\ref{sec:ss_design_for_pct}). The poor accuracy of MIS$_\text{con}$ is primarily caused by the requirement to design continuous models. The table suggests that the accuracy of MIS$_\text{con}$ is decreased with the SS input structure compared to its performance with the Ref input structure. The additional variable within the SS input structure appears to be unhelpful for MIS$_\text{con}$ accuracy, and it further increases the negative impact of the model continuity constraint on the MIS$_\text{con}$ performance.

The results further indicate that the optimized data labeling within the MIS$_\text{con,lab}$ approach significantly improves its accuracy, considering both input structures. The accuracy of MIS$_\text{con,lab}$ is not as high as that of MIS$_\text{SotA}$ but is comparable to that of SIS$_\text{Ref}$ and SIS$_\text{SS}$ in Table~\ref{tab:vgh_SIS_res}. Moreover, the MIS$_\text{con,lab}$ approach ensures continuity when switching between the designed models, which can be crucial in specific applications, particularly if the inferential sensor is part of a process control strategy. MIS$_\text{con,lab}$ achieves higher accuracy with the SS input structure than with the Ref input structure. Unlike the MIS$_\text{con}$ approach, the optimized data labeling enables the MIS$_\text{con,lab}$ approach to effectively use the additional variable within the SS input structure with respect to the resulting accuracy of MIS$_\text{con,lab}$.

\begin{figure}
	\centering
	\begin{subfigure}[b]{\textwidth}
		\centering
		\includegraphics[width=.9\linewidth]{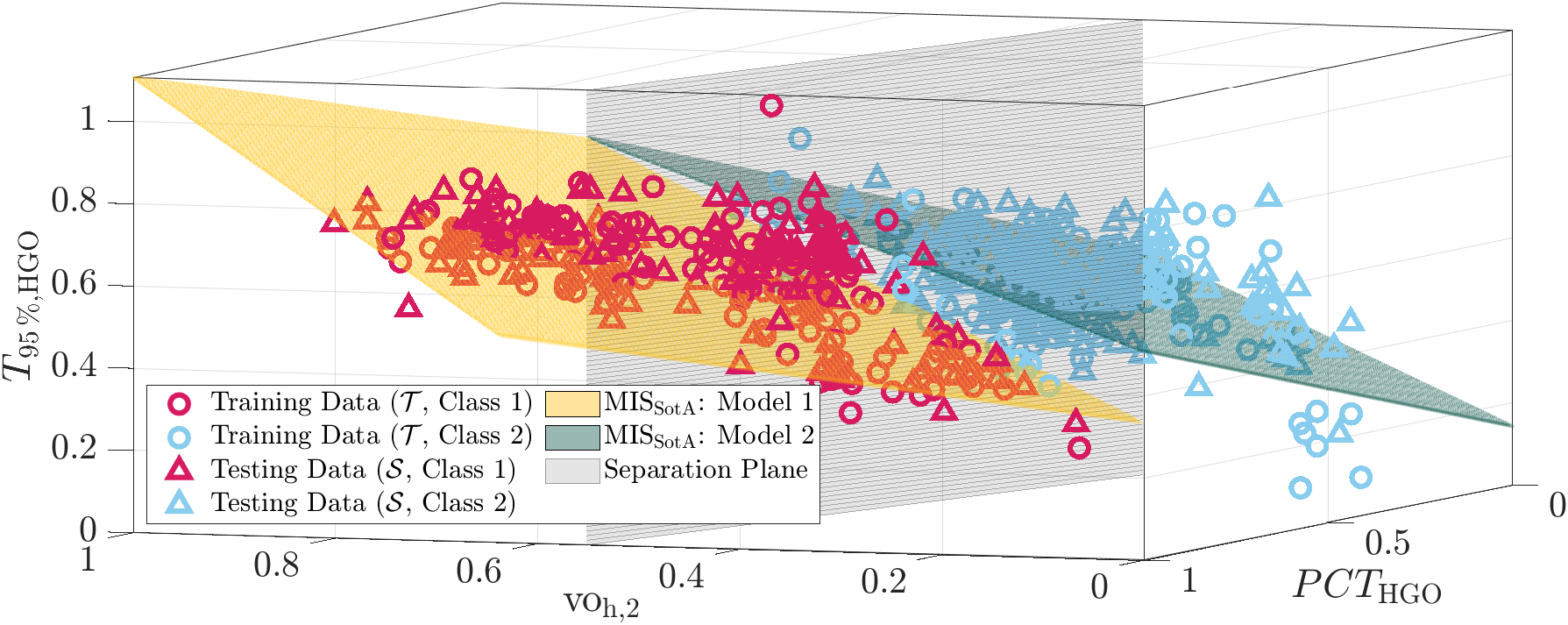}
		\caption{MIS$_\text{SotA}$: model 1 ($\text{RMSE}=[0.091(\mathcal T),0.091(\mathcal S)]$), model 2 ($\text{RMSE}=[0.102(\mathcal T),0.094(\mathcal S)]$).}
		\label{fig:vgh_MIS_SotA}
	\end{subfigure}\vspace{.2cm}
	\begin{subfigure}[b]{\textwidth}
		\centering
		\includegraphics[width=.9\linewidth]{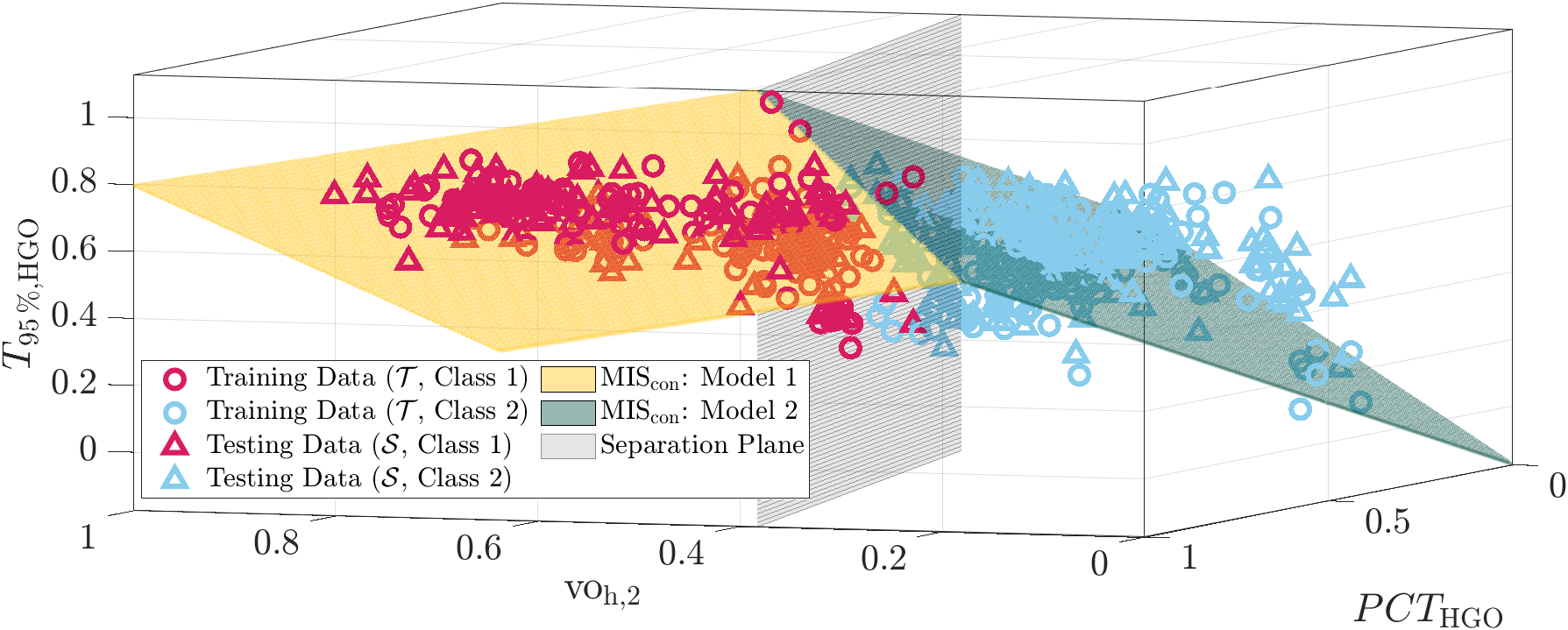}
		\caption{MIS$_\text{con}$: model 1 ($\text{RMSE}=[0.125(\mathcal T),0.133(\mathcal S)]$), model 2 ($\text{RMSE}=[0.136(\mathcal T),0.134(\mathcal S)]$).}
		\label{fig:vgh_MIS_con}
	\end{subfigure}\vspace{.2cm}
	\begin{subfigure}[b]{\textwidth}
		\centering
		\includegraphics[width=.9\linewidth]{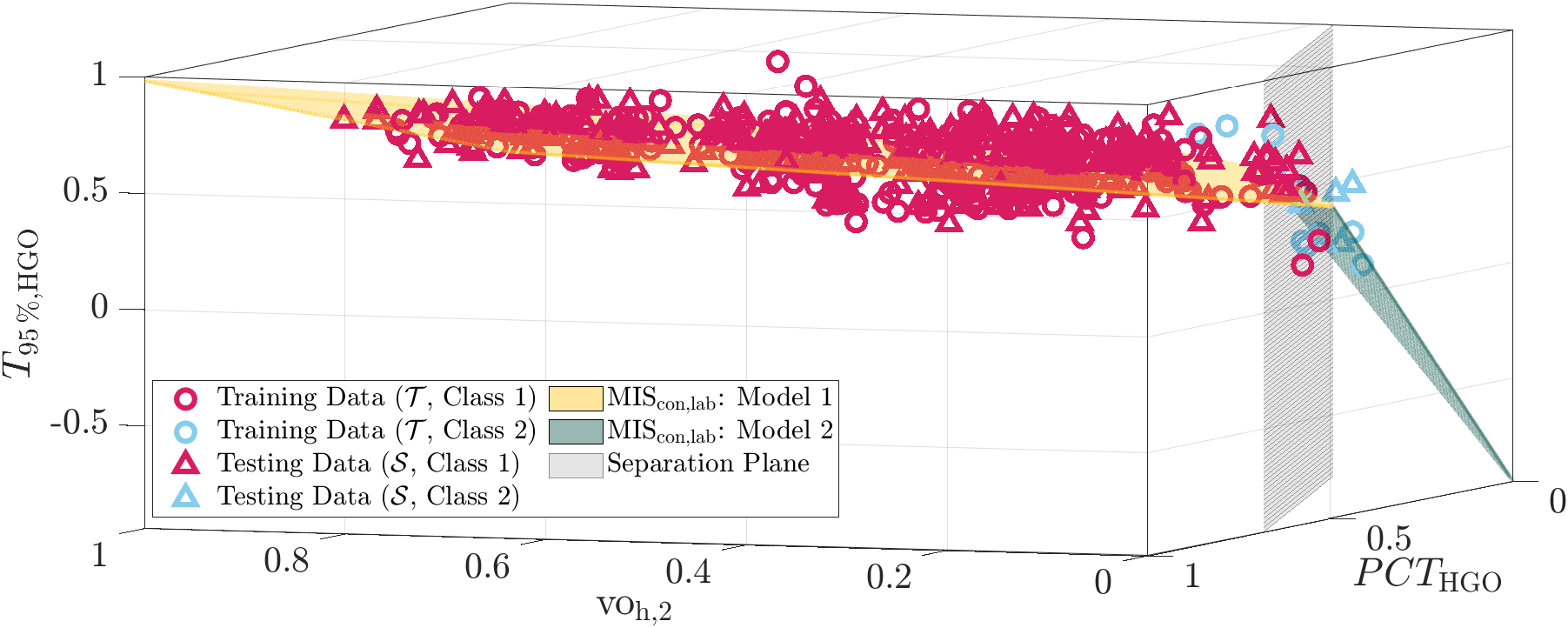}
		\caption{MIS$_\text{con,lab}$: model 1 ($\text{RMSE}=[0.099(\mathcal T),0.095(\mathcal S)]$), model 2 ($\text{RMSE}=[0.13(\mathcal T),0.211(\mathcal S)]$).}
		\label{fig:vgh_MIS_con_lab}
	\end{subfigure}
	\caption{The comparison of designed MISs on the industrial dataset from the VGH unit.}
	\label{fig:vgh_MIS}\vspace{0cm}
\end{figure} 

The designed MISs considering the SS input structure ($n_\text{p}^*=2$) are illustrated in Figure~\ref{fig:vgh_MIS}. The yellow and dark green surfaces represent the designed models within the MIS structure, and the gray vertical surface is the separation hyperplane. The circles represent the training dataset, and the triangles form the testing dataset. The pink color of circles or triangles represents measurements from the first class, and the blue color of circles or triangles indicates measurements from the second class.

The models of MIS$_\text{SotA}$ are shown in Figure~\ref{fig:vgh_MIS_SotA}. The model surfaces exhibit an obvious discontinuity at the interface (separation plane) between models. The discontinuity appears to be beneficial for model accuracy. The MIS$_\text{con}$ models are shown in Figure~\ref{fig:vgh_MIS_con}. We can see that the designed models are are continuous, which does not allow the arbitrary rotation of the MIS$_\text{con}$ model surfaces. Therefore, the models of MIS$_\text{con}$ deviate more from the measurements more than the models of MIS$_\text{SotA}$, resulting in lower accuracy of the MIS$_\text{con}$ model compared to MIS$_\text{SotA}$.

The designed models of MIS$_\text{con,lab}$ are shown in Figure~\ref{fig:vgh_MIS_con_lab}. We can observe the continuity at the switch between the MIS$_\text{con,lab}$ models. We can also see that one data class involves the majority of the measurements, while the remaining points are assigned to another, smaller class. This occurs when there are no discernible classes of measurements in the provided dataset (as seen in Figures~\ref{fig:pct_MIS_con_lab_rnd} and~\ref{fig:vgh_MIS_con_lab}), although this is not always the case (as seen in Figures~\ref{fig:pct_MIS_con_lab_2cls}). In the case of indistinct classes of measurements in the available dataset, MIS$_\text{con,lab}$ attempts to improve the accuracy of the model designed on the majority of measurements by assigning the most deviated measurements to the smaller class. As shown in Figure~\ref{fig:pct_MIS_con_lab_rnd}, the measurements from the smaller class have the potential to explain the nonlinear nature of the estimated variable. The results from MIS$_\text{con,lab}$ indicate that the studied industrial unit is operated primarily an operating range, where one model suffices to capture the behavior of the inferred output. The second model, on the other hand, involves measurements from other operating regimes, which occurs only occasionally in the plant operation. This observation leads us to conclude that the proposed MIS$_\text{con,lab}$ approach effectively captures the different operating conditions of the industrial unit.

\section{Discussion}\label{sec:discussion}
Multi-model linear inferential sensors offer several advantages over single-model (linear or nonlinear) sensors. They can enhance the model performance over the linear SISs without compromising on the model complexity (and implied usability) as nonlinear SISs do. As for any other data-based model, there is a risk of overfitting, which is connected to the number of chosen models within the MIS structure. In this contribution, we used the simplest MIS structure with two concurrent models involved. The presented methodology for the MIS design can be easily extended to involve more than two models. We have already designed MIS considering more than two models for several simulation cases. It seems that the increasing number of designed models can further increase the accuracy of MIS. On the other hand, this type of MIS has higher requirements for the quantity and quality of data compared to the MIS with two models. A practical strategy to decide on the number of models might use cross-validation as commonly exploited in overfit mitigation for SISs.

Additionally, MIS has the potential to reduce the number of input variables and thus simplify the sensor structure. The MIS design on the industrial dataset (see in Section~\ref{sec:ss_design_for_vgh}) considered only the reference input structure given by~\eqref{eq:vgh_SIS_ref} and the enhanced input structure determined by the subset selection approach exploring all possible structures with one or two input variables. In the case of the VGH unit, a simple input structure seems to be desired, which is confirmed by the excellent performance of SIS$_\text{Ref}$ considering only one input variable. Another possible approach to finding a suitable input structure for MIS is to combine cross-validation with some feature selection approach taking into account the objectives of the MIS design. An advanced alternative to determine the MIS input structure would involve extension of the objective function of MIS by an appropriate penalization element, whose purpose would be to reduce the absolute value of the model parameters. Such penalization is considered, e.g., in the LASSO (1-norm) or ridge (2-norm) regression approaches. The enhanced form of the MIS design would directly provide the optimal input structure for the designed models and would allow for a variation of input set among different individual models within the MIS. Establishment of continuity within such MIS is not straightforward and should be explored in the future works.

The presented case studies of the inferential sensor design provide several important insights about how to choose the appropriate MIS approach in a particular situation. The design of MIS$_\text{SotA}$ should be performed if the studied process requires an inferential sensor with high accuracy, reliable knowledge about different operating regimes is provided, and discontinuous in switching between the models cannot cause any (e.g., stability) issues within the considered process. If all previous specifications remain the same but continuity of the designed models is necessary, then MIS$_\text{con}$ should be considered. The results indicated that the continuity of the models is achieved at the expense of the inferential sensor accuracy. In cases when reliable knowledge about different operating regimes within the process is not provided, then MIS$_\text{con,lab}$ represents the best option. In the case that the previous specification remains but the continuity when switching is not necessary, it is possible to solve problem~\eqref{eq:mis_con_lab} with a relaxation of continuity constraints~\eqref{eq:mis_con_lab_const_4}.

\section{Conclusions}\label{sec:conclusions}
In this paper, novel approaches for multi-model inferential sensor (MIS) design were introduced. These approaches provide (a) continuous switching between models of the inferential sensor and (b) optimized data labeling of the training dataset. The performance of the studied MISs was compared against the performance of the single-model inferential sensors (SIS) on the datasets from two case studies from the petrochemical industry: (1) the model of pressure compensated temperature and (2) the Vacuum Gasoil Hydrogenation unit.

The statistical evaluation of the results from the design of the inferential sensor for the model of pressure compensated temperature shows that MISs outperform SISs, even the nonlinear ones. The results indicate that the performance of MISs is highly affected by the information content within the training dataset. If the data within the training dataset represent different operating regimes than the testing data, then MISs achieved similar performance compared to SIS. Moreover, the analysis of the impact of the output variable noise on the performance of the inferential sensors indicated that MIS achieved higher accuracy than SIS over the entire studied set of noise variances.

The design of inferential sensors on the industrial dataset from the Vacuum Gasoil Hydrogenation unit showed that the MIS design can provide the inferential sensor with higher accuracy compared to the SIS design. It seems that MIS can outperform the accuracy of the currently implemented reference SIS by about 6\,\% considering the same input structure or by about 8\,\% considering enhanced input structure.

\section*{Acknowledgments}\label{sec:acknowledgments}
We are grateful to the anonymous reviewers for their constructive comments that helped greatly improve the quality of the manuscript. This research is funded by the Slovak Research and Development Agency under the projects APVV-21-0019 and APVV SK-FR-2019-0004, by the Scientific Grant Agency of the Slovak Republic under the grants VEGA 1/0691/21 and VEGA 1/0297/22, and by the European Commission under the grant no. 101079342 (Fostering Opportunities Towards Slovak Excellence in Advanced Control for Smart Industries).

\bibliographystyle{elsarticle-harv}
\bibliography{other/references}
\end{document}